\documentclass[smallcondensed]{svjour3}

\usepackage[latin1]{inputenc}
\usepackage{epsfig}
\usepackage[]{amsmath}
\usepackage[]{amssymb}
\usepackage[]{amscd}
\usepackage[mathcal]{euscript}
\usepackage{amsfonts}
\usepackage{mathrsfs}
\usepackage{graphicx}
\usepackage{algo}
\usepackage{algorithm}
\usepackage{algorithmic}
\usepackage{boxedminipage}
\usepackage{bm,bbm}

\usepackage{natbib}

\bibpunct{(}{)}{;}{a}{,}{,}

\usepackage{colortbl}

\newtheorem{crit}[theorem]{Criterion}
\newtheorem{ass}[theorem]{Assumption}

\newcommand{\mse}{\text{MSE}}

\newcommand{\statespace}{{\mathbb{S}}}

\newcommand{\eP}{\mathbf{\bar P}}
\newcommand{\mat}[1]{\mathbf{#1}}
\newcommand{\prob}{\mathbbm{P}}
\newcommand{\expect}{\mathbbm{E}}
\newcommand{\var}{\mathbbm{V}}

\newcommand{\sS}{\mathcal{S}}
\newcommand{\sT}{\mathcal{T}}

\newcommand{\vx}{\mathbf{x}}
\newcommand{\lp}[1]{$L^#1$}

\journalname{ARXIV Version}
\date{6.6.2009}

\begin{document}
\title{The Optimal Unbiased Value Estimator and its Relation to LSTD, TD and MC}
\titlerunning{Optimal Unbiased Value Estimation}

\author{Steffen Gr\"unew\"alder \and Klaus Obermayer}
\institute{S. Gr\"unew\"alder \and K. Obermayer \at  Department of Computer Science, 
       Berlin University of Technology, 
       Berlin, 10587, Germany 
\and 
S. Gr\"unew\"alder \at Centre for Computational 
Statistics and Machine Learning,
University College London, Gower Street,
London WC1E 6BT, United Kingdom \\
\email{steffen@cs.ucl.ac.uk}, Tel: 0044 (0)20 7679 0384, Fax: 0044 (0)20 7387 1397}

\maketitle

\begin{abstract}%   
In this analytical study we derive the optimal unbiased value estimator (MVU)
and compare its statistical risk to three well known value estimators: 
Temporal Difference learning (TD), Monte Carlo estimation (MC) and Least-Squares Temporal Difference Learning (LSTD).
We demonstrate that LSTD is equivalent to the MVU if the Markov Reward Process (MRP) is acyclic
and show that both differ for most cyclic MRPs as LSTD is then typically biased. More generally, 
we show that estimators that fulfill the Bellman equation can only be unbiased for special cyclic MRPs.
The main reason being the probability measures with which the expectations are taken. These measure vary 
from state to state and due to the strong coupling by the Bellman equation it is typically not possible for a 
set of value estimators to be unbiased with respect to each of these measures.
Furthermore, we derive relations of the MVU to MC and TD. The most important one being 
the equivalence of MC to the MVU and to LSTD for undiscounted MRPs in which MC has the 
\textit{same amount of information}. In the discounted case this equivalence does not hold anymore.
For TD we show that it is essentially unbiased for acyclic MRPs and biased for cyclic MRPs. 
We also order estimators according to their risk and present counter-examples to show that no general ordering 
exists between the MVU and LSTD, between MC and LSTD and between TD and MC. 
Theoretical results are supported by examples and an empirical evaluation.
\end{abstract}

\keywords{
Optimal Unbiased Value Estimator \and  Maximum Likelihood Value Estimator \and
Sufficient Statistics \and Rao-Blackwell Theorem}

\section{Introduction}
One of the important theoretical issues in reinforcement learning are rigorous statements on
convergence properties of so called \textit{value estimators}
(e.g. \citep{SUTT88}, \citep{DAY92}, \citep{JAAK94}, \citep{BRAD96})
which provide an empirical estimate of the expected future reward for every given state. 
So far most of these convergence results  were restricted to the asymptotic case 
and did not provide statements 
for the case of a finite number of observations. In practice, however, one
wants to choose the estimator which yields the best result for a 
given number of examples or in the shortest time.

Current approaches to the finite example case are mostly
empirical and few non-empirical approaches exist.
\citep{KEAR00} present upper bounds on the generalization error for \textit{Temporal Difference estimators (TD)}.
They use these bounds to formally verify the intuition that TD methods are subject to a ``bias-variance'' trade-off
and to derive ``schedules'' for estimator parameters. Comparisons of different estimators with respect to the bounds were not performed.
The issue of  \textit{bias and variance} in reinforcement learning is also addressed in other works
(\citep{SING98}, \citep{MAN07}). 
\citep{SING98} provide analytical expressions of the \textit{mean squared error (MSE)} for various \textit{Monte Carlo (MC)} and 
TD value estimators. Furthermore, they provide a software that yields the exact mean squared error curves given a complete description 
of a \textit{Markov Reward Process (MRP)}.
The method can be used to
compare different estimators for concrete MRPs.
But it is not possible to prove general statements with their method.
The most relevant works for our analysis are provided by  \citep{MAN07} and by \citep{SING96}.

In \citep{MAN07} the bias and the variance in value function estimates is studied and 
closed-form approximations are provided for these terms. The approximations are used in 
a large sample approach to derive asymptotic confidence intervals. The underlying
assumption of normally distributed estimates is tested empirically on a dataset of
a mail-order catalogue. In particular, a Kolmogorov-Smirnov test was unable to reject 
the hypothesis of normal distribution with a confidence of 0.05. The value function
estimates are based on sample mean estimates of the MRP parameters. The parameter estimates
are used in combination with the value equation to produce the value estimate. Different assumptions
are made in the paper to simplify the analysis. A particularly important 
assumption is that the number of visits to a state is fixed.  
Under this assumption the sample mean parameter estimates
are unbiased and the application of the value equation results in biased estimates. We show that without this 
assumption the sample mean estimates underestimate the parameters in the average and the value estimates can
therefore be unbiased in special cases. We address this point in detail in Section \ref{LSTD_Sec}.

In \citep{SING96} different kinds of eligibility traces are introduced and analyzed. It is shown that
TD($1$) is unbiased if the \textit{replace-trace} is used and that it is biased if the usual eligibility trace 
is used. What is particularly important for our work is one of their side findings: The Maximum Likelihood and 
the MC estimates are equivalent in a special case. We characterize this special case with 
Criterion \ref{crit:full_inf} (p. \pageref{crit:full_inf}) and we make frequent use of this property. 
We call the criterion the \textit{Full Information Criterion} because all paths that are relevant for a 
value estimator in a state $s$ must hit this state (For details see p. \pageref{crit:full_inf}). 

\begin{figure}[h!]
\begin{flushleft}
\setlength{\epsfxsize}{4.7in}
\epsfbox{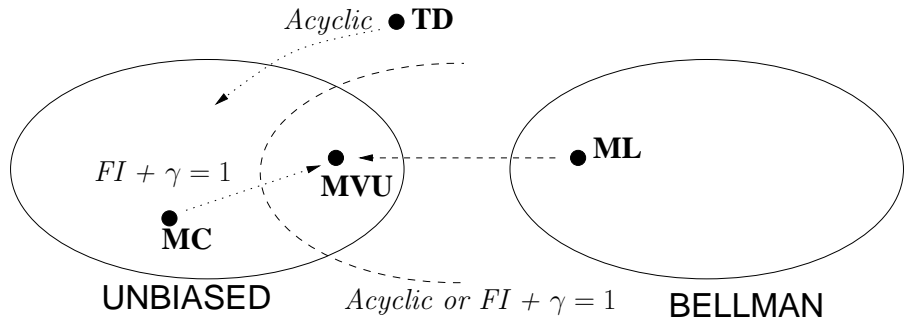} 
\caption{The figure shows two value estimator classes and four value estimators. On the left the 
class of unbiased value estimators is shown and on the right the class of Bellman estimators.  The graph visualises
to which classes the estimators belong and how the two classes are related. 
The cursive texts state conditions under which different estimators 
are equivalent, respectively, under which the two classes overlap. \textit{FI} denotes the Full Information
Criterion.
}
\label{Overview}
\vspace{-0.5cm}
\end{flushleft}
\end{figure}

In this paper we follow a new approach to the finite example case using tools from statistical estimation theory (e.g. 
\citep{KEND91}). Rather than relying on bounds, on approximations, or on results to be recalculated 
for every specific MRP this 
approach allows us to derive 
general statements.
Our main results are sketched in Figure \ref{Overview}. The major contribution is the derivation of
the optimal unbiased value estimator (\textit{Minimum Variance Unbiased estimator (MVU)}, 
Sec. \ref{sec:MVU}). We show that the \textit{Least-Squares Temporal Difference estimator (LSTD)} 
from \citep{BRAD96} is equivalent to the \textit{Maximum Likelihood value estimator (ML)}  
(Sec. \ref{EquLSTD}) and that both are equivalent to the MVU if the discount $\gamma=1$ (undiscounted) 
and the Full Information Criterion 
is fulfilled or if an acyclic MRP is given (Sec. \ref{LSTD_Opt}). 
In general the ML estimator differs from the MVU because ML fulfills the Bellman equation and
because estimators that fulfill the Bellman equation can in general not be unbiased 
(We refer to estimators that fulfill the Bellman equation in the future as \textit{Bellman estimators}). 
The main reason for this effect being the probability measures with which the expectations 
are taken (Sec. \ref{sec:Unbiased_Bellman}). 
The bias of the Bellman estimators vanishes exponentially in 
the number of observed paths. As both estimators differ in general it is natural to ask which of them 
is better? We show that in general neither the ML nor the MVU estimator are superior 
to each other, i.e. examples exist where the MVU is superior and examples exist where ML is superior
 (Appendix \ref{sec:counter_mvu_ml}).

The \textit{first-visit MC} estimator is unbiased \citep{SING96} and therefore inferior to the MVU. 
However, we show that for $\gamma = 1$  the estimator becomes equivalent 
to the MVU if the Full Information Criterion  applies (Sec. \ref{MCRelSec}). Furthermore, we
show that this equivalence is restricted to the undiscounted case. 

Finally, we compare the estimators to TD($\lambda$). We show that TD($\lambda$) is essentially unbiased for acyclic MRPs
(Appendix \ref{sec:unbiased-td0-version}) and is thus inferior to the MVU and to the ML estimator for this case.
In the cyclic case TD is biased (Sec. \ref{sec:appr-equal-weight}). 

An early version of this work was presented in \citep{GRUEN07}. The analysis was restricted to acyclic MRPs and
to the MC and LSTD estimator. The two main findings were that LSTD is unbiased and optimal for acyclic MRPs and that 
MC equals LSTD in the acyclic case if the Full Information Criterion applies and $\gamma = 1$. It turned out that the 
second finding was already shown in more generality by \citep{SING96}[Theorem 5]. The restriction to acyclic MRPs 
simplified the analysis considerably compared to the general case which we approach in this work.
\quad \\
Theoretical findings are summarized in two tables in section \ref{sec:summary_theory} (p. \pageref{sec:summary_theory}).
Symbols are explained at their first occurrence and a table of notations is included in Appendix \ref{sec:notation}. 
For the sake of readability proofs are presented in Appendix \ref{sec:proofs}.

\section{Estimation in Reinforcement Learning}
\label{sec:estim-reinf-learn}
A common approach to the optimization of a control policy is to iterate between estimating the current
performance (value estimation) and updating the policy based on this estimate (policy improvement).
Such an approach to optimization is called \textit{policy iteration} \citep{SUTT98,BERT96}. The value
estimation part is of central importance as it determines the direction of the policy improvement step. 

In this work we focus on this value estimation problem and we study it for Markov Reward Processes.
In Reinforcement Learning Markov Decision Processes are typically used. A MRP is the same with
the only difference being that the policy does not change over time.

\subsection{Markov Reward Processes}
A Markov Reward Process consists of a state space $\mathbb{S}$ (in 
our case  a finite state space), probabilities $p_i$ to start in state $i$, 
transition probabilities $p_{ij}$ and a random reward $R_{ij}$ between 
states $i$ and $j$. The MRP is \textit{acyclic} if no state $i$ and no path 
$\pi = (s_1, s_2, s_3, \ldots)$  
exists such that $P(\pi) := p_{s_1s_2} p_{s_2s_3} \ldots > 0$ and state 
$i$ is included at least twice in $\pi$. 

Our goal is to estimate the values $V_i$ of the states in $\mathbb{S}$, i.e.
the expected future reward received after visiting state $i$. The value 
is defined as
\begin{align*}
&V_i = \sum_{j \in \mathbb{S}} p_{ij} \bigl(\mathbbm{E}[R_{ij}] + \gamma V_{j}\bigr) \text{ and  in vector notation by }
\mathbf{V} = \sum_{t=0}^\infty \gamma^t \mathbf{P}^t \mathbf{r} = (\mathbf{I} - \gamma \mathbf{P})^{-1} \mathbf{r}, 
\end{align*}
where $\mathbf{P}=(p_{ij})$ is the transition matrix of the Markov process, $\mathbf{I}$ the identity matrix,
 $\gamma \in (0,1]$  a discount factor and $\mathbf{r}$ is the vector of the expected one step reward 
($\mathbf{r}_i = \sum_{j \in \statespace} p_{ij}\expect[R_{ij}]$). In the undiscounted case ($\gamma =1$) we assume
that with probability one a path reaches a terminal state after a finite number of steps.

A large part of this work is concerned with the relation between the maximum likelihood value estimator
and the optimal unbiased value estimator. In particular, we are interested in equivalence statements for these 
two estimators. Equivalence between these estimators can only hold if the estimates for the reward are equivalent,
meaning that the maximum likelihood estimator for the reward distribution matches with the optimal unbiased estimator.
We therefore restrict our analysis to reward distributions with this property, i.e. we assume throughout that
the following assumption holds:
\begin{ass} 
The maximum likelihood estimate of the mean reward is unbiased and equivalent to
the optimal unbiased estimate.
\end{ass}
The assumption is certainly fulfilled for \textit{deterministic rewards}. Other important cases  
are \textit{normal distributed,  binomial and multinomial distributed
rewards}.

\subsection{Value Estimators and Statistical Risk}
We compare value estimators with respect to their risk (not the empirical risk)
$$\expect[\mathcal{L}(\bar V_i,V_i)],$$ 
where $\bar V_i$ is a value estimator 
of state $i$ and $\mathcal{L}$ is a loss function, which penalizes the deviation 
from the true value $V_i$.
We will mainly use the mean squared error
\begin{equation} \label{eq:MSE}
\mse[\bar V_i] := \expect[(\bar V_i-V_i)^2], 
\end{equation}
which can be split into a \textit{bias} and
a \textit{variance} term
$$ \mse[\bar V_i] = \underbrace{\var[\bar V_i]}_{\text{Variance}} 
+ (\underbrace{\expect[\bar V_i - V_i]}_{\text{Bias}})^2.$$
An estimator is called \textit{unbiased} if the bias term is zero. 
The unbiasedness of an estimator depends on the underlying probability distribution with which the mean is calculated.

Typically, there is a chance that a state is not visited at all by an agent and it makes no sense
to estimate the value if this event occurs. We encode the probability event that state $i$ has not 
been visited with $\{N_i = 0\}$ and that is has been visited at least once with $\{N_i \geq 1\}$, where 
$N_i$ denotes the number of visits of state $i$. Unbiased estimators are estimators that are correct in the mean.
However, if we take the (unconditional) mean for a MRP then we include the term $\expect[\bar V_i |\{N_i=0\}]$ into
the calculation, i.e. the value estimate for the case that the estimator has not seen a single example. This is 
certainly not what we want. We therefore measure the bias of an estimator using the conditional expectation 
$\expect[\thinspace \cdot \thinspace \thinspace | \{N_i \geq 1\}]$.

\paragraph{Equal Weighting of Examples}
\label{ExampleWeight}
We conclude this section by citing a simple criterion with which 
it is possible to verify unbiasedness and minimal MSE in special cases. 
This criterion provides an intuitive interpretation of a weakness of the TD($\lambda$) estimator
(see Section \ref{sec:appr-equal-weight}).
Let $x_i, i=1, \ldots, n$ 
be a sample consisting of $n\geq 1$ 
\textit{independent and identically distributed (iid)} elements of an arbitrary distribution.
The estimator
\begin{equation} 
\label{eq:lin_est}
\sum_{i=1}^n \alpha_i x_i \text{, \quad with \quad } 0 \leq \alpha_i \leq 1\text{, \quad and \quad }  \sum_{i=1}^n 
\alpha_i = 1,
\end{equation}
is unbiased and has the lowest variance for $\alpha_i = 1/n$ \citep{KEND91}. 
The $x_i$ could, for example, be the summed rewards for $n$ different paths starting in the same state $s$, i.e. 
$x_i:= \sum_{t=0}^\infty \gamma^t R_t^{(i)}$, where $R_t^{(i)}$ denotes the reward at time $t$ in path $i$.
The criterion states that for estimators which are linear combinations 
of iid examples all examples should have an equal influence and none should be preferred over another. 
However, it is important to notice that not all unbiased estimators must be linear combinations of such sequences and
that better unbiased estimators might exist. In fact this is the case for MRPs. The structure of a MRP allows better
value estimates.

\subsection{Temporal Difference Learning}
A commonly used value estimator for MRPs is the TD($\lambda$) estimator \citep{SUTT88}. 
It converges on average ($L^1$-convergence, \citep{SUTT88}) and 
it converges almost surely to the correct value
\citep{DAY92,JAAK94}.
In practical tasks it seems to outperform the MC estimator with respect to convergence speed 
and its computational costs are low. Analyses for the TD(0) estimator are often 
less technical. We therefore restrict some statements to this estimator. TD(0) can be defined by means of an
update equation:
\begin{equation}
\label{TD0Def}
\bar V^{(i+1)}_s = \bar V^{(i)}_s + \alpha_{i+1} ( R_{ss'}^{(i+1)} + \gamma \bar V^{(i)}_{s'} - \bar V^{(i)}_s),
\end{equation}
where $\alpha_{i+1}$ is the learning rate, $\bar V^{(i)}_s$ is the estimated value for state $s$ after the $i$th 
transition, 
$s'$ is the successor state of $s$  and $R_{ss'}^{(i+1)}$ is 
the reward which occurred during the transition from $s$ to $s'$. 
The general TD($\lambda$) update equation is given by
\begin{gather*}
\bar V_s^{(i+1)} = \bar V_s^{(i)}  + \Delta \bar V_s^{(i+1)}  \text{\quad and \quad } 
\Delta \bar  V_s^{(i+1)} = \alpha_{i+1} (R_{ss'}^{(i+1)}  + \gamma \bar V_{s'}^{(i)}  - \bar V_s^i)  e_s^{(i+1)}, 
\end{gather*}
where $\alpha_i$ is the learning rate in sample path $i$ (the learning rate might be defined differently) and
$e_s^{(i+1)}$ is an \textit{eligibility trace}.
The update equation can be applied 
after each transition (\textit{online}), when a terminal state is reached (\textit{offline}) 
or after an entire set of paths has been observed (\textit{batch update}). 
The eligibility trace can be defined
in various ways. Two important definitions are the \textit{accumulating trace} and the \textit{replacing trace}
\citep{SING96}. 
In \citep{SING96} it is shown that for $\lambda = 1$ the TD($\lambda$) estimator corresponding to the 
 accumulating trace is biased while the one corresponding to the  replacing trace is unbiased.
The replacing trace is defined by
\begin{equation}
e_{s}^{(i+1)} = 
\begin{cases}
1  & \text{ if } s= t, \\
\gamma \lambda & \text{ else. }
\end{cases}
\end{equation}
For acyclic MRPs both definitions are equivalent. For $\lambda < 1$ the 
estimators are biased  towards their initialization value. 
However, a minor modification is sufficient to delete the bias for acyclic MRPs (App. \ref{sec:unbiased-td0-version}
on p. \pageref{sec:unbiased-td0-version}). We will mostly use this modified version.

\subsection{Monte Carlo Estimation}
The Monte Carlo estimator is the sample mean estimator of the summed future reward \citep{SUTT98}. 
For acyclic MRPs the MC estimator is given by
$$\frac{1}{n} \sum_{i=1}^n  \left( \sum_{t=0}^\infty \gamma^t R_t^{(i)} \right),$$ 
where $n$ is the number of paths that have been observed. 

In the cyclic case there are two alternative MC estimators: \textit{First-visit MC} and \textit{every-visit MC}. 
First-visit MC makes exactly one update for each visited state. It uses the part of the path which follows upon 
the first visit of the relevant state. The first-visit MC estimator $\bar V_i$ is unbiased for every state 
$i$, i.e. $\expect[\bar V_i|{N_i \geq 1}]=V_i$. Every-visit MC makes an update for
each visit of the state. The advantage of the every-visit MC estimator is that it has 
more samples available for estimation, however, the paths overlap and the estimator is therefore biased 
\citep{SING96}. Both estimators converge almost surely and on average to the correct value. 

The MC estimators are special cases of TD($\lambda$). The every-visit MC estimator is equivalent to 
TD($\lambda$) for the \textit{accumulate trace} and the 
first-visit MC estimator for the \textit{replace trace} if $\lambda=1$ and $\alpha_i = 1/i$.

\section{Comparison of Estimators: Theory}
\label{sec:estimators-which-use}
The central theme of this paper is the relation between two important classes of value estimators and between four 
concrete value estimators. One can argue that the two most important estimator classes are the 
estimators that \textit{fulfill the Bellman equation} and estimators that are \textit{unbiased}. 
The former class is certainly of great importance as the Bellman equation is the central equation in Reinforcement 
Learning. The latter class proved its importance in statistical estimation theory, where it is the central 
class of estimators that is studied. We analyse the relation between these two classes.

On the estimator side we concentrate on popular Reinforcement Learning estimators (the Monte-Carlo and the 
Temporal Difference estimator) and on estimators that are optimal in the two classes.
These are: (1) The optimal unbiased  value estimator which we derive in Section \ref{sec:Unbiased_Bellman}.
(2) The Maximum Likelihood (ML) estimator for which one can argue (yet not prove!) 
that it is the best estimator in the class of Bellman estimators.

Parts of this section are very technical. We therefore conclude this motivation with a high level overview 
of the main results.

\paragraph{Estimator Classes: Unbiased vs. Bellman Estimators}
The key finding for these two estimator classes is that cycles in an MRP essentially separate them.
That means if we have a MRP with cycles then the estimators can either fulfill the Bellman equation or 
at least some of the value estimators must be biased. The main factor that is responsible for this effect is
the ``normalization'' $\{N_i \geq 1\}$. The Bellman equation couples the estimators, yet the estimators must be ``flexible'' to be
unbiased with respect to different probability measures, i.e. the conditional probabilities 
$\prob[\thinspace \cdot \thinspace \thinspace |\{N_i \geq 1\}]$.

Furthermore, we show that the discount has an effect onto the bias of Bellman estimators. 
Estimators that use the Bellman equation are based on parameter estimates $\bar p_{ij}$. We show that these
parameter estimates must be discount dependent. Otherwise, a ``further bias'' is introduced.

We show that these factors are the main factors for the separation of the classes: (1) If the MRP is acyclic 
or (2) if the problem with the normalization and the discount is not present then Bellman estimators 
can be unbiased. 

\paragraph{Estimator Comparison and Ordering: MVU, ML, TD and MC}
The key contribution in this part is the derivation of the optimal unbiased value estimator.
We derive this estimator by conditioning the first-visit Monte Carlo estimator with ``all the information'' 
that is available through the observed paths and we show that the resulting estimator is optimal 
with respect to any convex loss function. The conditioning has two effects: (1) The new estimator uses the Markov 
structure to make use of (nearly) all paths. (2) It uses ``consistent'' alternative cycles beside the observed ones. 
For example, if a cyclic connection from state $1 \rightarrow 1$ is observed once in the first run and three 
times in the second run, then the optimal estimator will use paths with the cyclic connection being taken
$0$ to $4$ times. Consistent with this finding, we show that if the first-visit MC estimator observes all paths  
and the modification of cycles has no effect, then the first-visit MC estimator is already optimal.  

Furthermore, the methods from statistical estimation theory allow us to establish a strong relation between 
the MVU and the Maximum Likelihood estimator. The ML estimator uses also all information, but it is 
typically biased as it fulfills the Bellman equation. However, in the cases where the ML estimator is unbiased
it is equivalent to the MVU. In particular, the ML estimator is unbiased and equivalent to the MVU for acyclic MRPs 
and for MRPs where the Full Information Criterion applies.

In the final theory part we are addressing the Temporal Difference estimator. 
In contrast to MC and ML the theoretical results for TD are not as strong. The reason being that the tools 
from statistical estimation theory that we are applying can be used to compare estimators inside one of the two 
estimator classes. However, TD is typically neither contained in the class of unbiased estimators nor in the 
class of Bellman estimators. We are therefore falling back to a more direct comparison of TD to ML. The analysis 
makes concrete the relation of the optimal value estimator to TD and demonstrates the powerfulness of the 
Rao-Blackwell theorem.

Beside the mentioned equivalence statements between different estimators we are also establishing orderings 
like ``the MVU is at least as good as the first-visit MC estimator'' or we are giving counter-examples if no 
ordering exists.

\subsection{Unbiased Estimators and the Bellman Equation}
\label{sec:Unbiased_Bellman}
In this section we analyze the relation between unbiased estimators and Bellman estimators.
Intuitively, we mean by ``a value estimator 
$\mathbf{\bar V}$ fulfills the Bellman equation'' that $\mathbf{\bar V} = \mathbf{\bar r} + \gamma \mathbf{\bar P} 
\mathbf{\bar V}$, where $\mathbf{\bar r}, \mathbf{\bar P}$ are the rewards, respectively the transition matrix, 
of a well defined MRP. We make this precise with the following definition:
\begin{definition}[Bellman Equation for Value Estimators.] 
An estimator $\mathbf{\bar V}$ fulfills the Bellman equation if
a MRP $\mathbf{\bar M}$ exists with the same state space as the original MRP, with 
a transition matrix $\mathbf{\bar P}$, deterministic rewards $\mathbf{\bar r}$ and 
with value $\mathbf{\bar V}$, i.e. $\mathbf{\bar V} = \mathbf{\bar r} + \gamma \mathbf{\bar P} \mathbf{\bar V}$.
Furthermore, $\mathbf{\bar M}$ is not allowed to have additional connections, i.e.  $\mathbf{\bar P}_{ij} = 0$ 
  if in the original MRP $\mathbf{ P}_{ij} = 0$ holds.
\end{definition}
Two remarks: Firstly, we restrict the MRP $\mathbf{\bar M}$ to have deterministic rewards for simplicity. 
Secondly, the last condition is used to enforce that the MRP $\mathbf{\bar M}$ has a ``similar structure'' 
as the original MRP. However, it is possible for $\mathbf{ \bar M}$ to have fewer connections. For example, this will
 be the case if not every transition $i\rightarrow j$ has been observed.

Constraining the estimator to fulfill the  Bellman equation restricts the class of estimators considerably. 
Essentially, the only degree of freedom is the parameter estimate $\mathbf{\bar P}$. If 
$\mathbf{I} - \gamma \mathbf{\bar P}$ is invertible then 
$$ \mathbf{\bar V} = (\mathbf{I} - \gamma \mathbf{\bar P})^{-1} \mathbf{\bar{r}} =: V(\mathbf{\bar P},\mathbf{\bar r}),$$
i.e. $\mathbf{\bar V}$ is completely specified by $\mathbf{\bar P}$ and $\mathbf{\bar r}$. Here, 
$V(\mathbf{\bar P},\mathbf{\bar r})$ denotes the value function for a MRP with parameters  
$\mathbf{\bar P}$ and  rewards $\mathbf{\bar r}$. In particular, the Bellman equation couples the value estimates
of different states. This coupling of the value estimates introduces a bias. The intuitive explanation of the bias is
the following: Assume we have two  value estimators $\bar V_i, \bar V_j$ and both are connected with a
connection $i \rightarrow j$ and $\bar p_{ij}=1$ holds. Fixing, $\expect[\bar V_j| \{N_j \geq 1\}] = V_j$ defines
then essentially the value for $\bar V_i$ as $\bar V_i = r_{ij} + \gamma \bar V_j$. Yet, the value for 
$\bar V_i$ must be flexible to allow $\bar V_i$ to depend on the probability of $\{ N_j \geq 1 \}$,
as $\expect[\bar V_i| \{N_i \geq 1\}] = V_i$  must hold. It is in general not possible to fulfill both 
constraints simultaneously in the cyclic case, i.e. constraining $\bar V_i$ for all
states $i$ and enforcing the Bellman equation. However, value estimators for single states can be unbiased, 
even if the Bellman equation is fulfilled.

Another factor that influences the bias is the discount $\gamma$. If the Bellman equation is fulfilled by 
$\mathbf{\bar V}$ then the value estimate can be written as $\sum_{t=0}^\infty \gamma^t 
\mathbf{\bar P}^t \mathbf{\bar r}$, i.e. $\gamma^t$ weights 
the estimate $\mathbf{\bar P}^t $ of $\mathbf{P}^t$. If $\expect[\mathbf{\bar P}^t] \not = \mathbf{P}^t$ and
the parameter estimate $\mathbf{\bar P}$ is independent of $\gamma$ then with varying $\gamma$ 
the deviations of $\mathbf{\bar P}^t $ from $\mathbf{P}^t$ are weighted differently and it
is intuitive that we can find a $\gamma$ for which the weighted deviations does not cancel out and
the estimator is not unbiased. This effect can be circumvented by making the parameter estimator
$\mathbf{\bar P}$ discount dependent.

\subsubsection{Normalization $\prob[\{N_i \geq 1\}]$ and Value Estimates on $\{N_i = 0\}$}
\label{sec:norm_prob}
Consider the MRP shown in Figure \ref{fig:cyclic_example} (B) and let the number of observed paths be one ($n=1$). 
The agent starts in state $2$ 
and has a chance of $p$ to move on to state $1$. 
The value of state $1$ and $2$ is 
  $$ V_1 = V_2 = (1-p) \sum_{i=0}^\infty i p^i = \frac{p}{1-p}. $$
Using the sample mean parameter estimate $\bar p = i/(i+1) $, we get the following value 
estimate for state 2:
  \begin{gather*}
    \bar V_2(i) = \frac{\bar p}{1-\bar p} =  i  \text{\quad } \Rightarrow \text{\quad}
    \expect[\bar V_2(i)] = (1-p) \sum_{i=0}^\infty \bar V_2(i) p^i = V_2, 
  \end{gather*}
where $\bar V_2(i)$ denotes the value estimate, given the cyclic transition has been taken $i$ times.
The estimator fulfills the Bellman equation. Therefore,  $\bar V_1(i) = \bar V_2(i) = i$, given at least one visit of
state $1$, i.e.  conditional on the event $\{N_1 \geq 1\}$.  The expected value estimate for state $1$ is therefore
\begin{gather*}
\expect[\bar V_1(i)|\{N_1\geq 1\}] = \frac{(1-p) \sum_{i=1}^\infty i p^i }{(1-p) \sum_{i=1}^\infty p^i}
= \frac{p/(1-p) }{1 - (1-p)}  = \frac{V_1}{p},
\end{gather*}
where $(1-p) \sum_{i=1}^\infty p^i=p$ is the normalization. Hence, the estimator is biased.

Intuitively the reasons for the bias are:
Firstly, $\bar V_1$ equals $\bar V_2$ on $\{N_1 \geq 1\}$ but the estimators differ (in general) on $\{N_1 = 0\}$. 
In the example, we made no use of this point. We could make use of it by introducing a  reward for the 
transition $2\rightarrow 3$. Secondly,
the normalization differs, i.e. $\expect[\thinspace \cdot \thinspace]$ versus 
$\expect[\thinspace \cdot \thinspace \thinspace |\{N_1\geq 1\}]$. In our example
we used this point. Both estimators are $0$ on $\{N_1=0\}$ and are therefore always equivalent. However, the
expectation is calculated differently and introduces the bias. 

The following Lemma shows that this problem does not depend on the parameter estimate we used:
\begin{lemma}[p. \pageref{proof:NormalizationBiased}]
\label{lem:NormalizationBiased}
For the MRP from Figure \ref{fig:cyclic_example} (B) there exists no parameter estimator $\bar p$
such that $V_i(\bar p)$ is unbiased for all states $i$.
\end{lemma}

How do these effects behave in dependence of the number $n$ of observed paths? Let $p_i$ denote the probability to
visit state $i$ in one sampled path. Then the probability of the event $\{N_i=0\}$ drops exponentially fast, i.e. 
$\prob[\{N_i=0\}]\leq (1-p_i)^n$ and the normalization $1/\prob[\{N_i\geq1\}]$ approaches one exponentially fast.
Therefore, if the estimates are upper bounded on $\{N_i= 0\}$  then the bias drops exponentially fast in $n$.

\subsubsection{Discount}
\begin{figure}[t]
\begin{center}
\setlength{\epsfxsize}{4.in}
\centerline{ \epsfbox{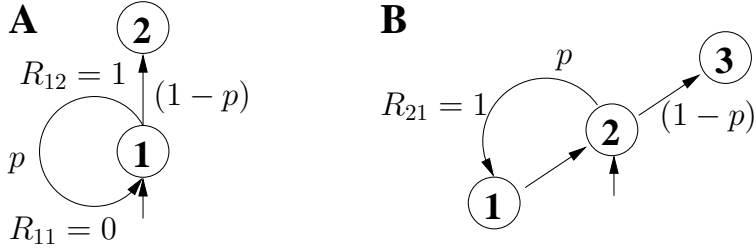}}
\caption{\textbf{A}: A cyclic MRP with starting state 1 and with probability p
for the cyclic transition. The reward is 1 for the cyclic transition and 0 otherwise.
\textbf{B}: A cyclic MRP with starting state 2 and with probability p
for the cyclic transition. The reward is 1 for the cyclic transition from state 2 to state 1 and 0 otherwise.
}
\label{fig:cyclic_example}
\end{center}
\end{figure}

Consider the MRP from Figure \ref{fig:cyclic_example} (A) for one run ($n=1$) and for $\gamma < 1$. We use again the sample mean parameter estimate, i.e. $\bar p = i/(i+1)$ if the cyclic transition has been taken $i$ 
times. 
The value of state $1$ is 
$$ V_1 = (1-p) \sum_{i=0}^\infty \gamma^i p^i = \frac{1-p}{1-\gamma p} \text{\quad and the value estimate is
 \quad }
 \bar V_1 = \frac{1 - i/(i+1)}{1- \gamma i/(i+1)}. $$
The  estimator is unbiased if and only if
\begin{align*}
&(1-p) \sum_{i=0}^\infty \gamma^i p^i \overset{?}{=} \expect[\bar V_1] = (1-p) \sum_{i=0}^\infty 
\frac{1 - i/(i+1)}{1- \gamma i/(i+1)} p^i. 
\end{align*}
The equality marked with $\overset{?}{=}$ holds if and only if
\begin{align*}
\sum_{i=0}^\infty \left(\gamma^i -
\frac{1 - i/(i+1)}{1- \gamma i/(i+1)}\right) p^i  = 0.
\end{align*}
With induction one sees that  $\gamma^i \leq \frac{1 - i/(i+1)}{1- \gamma i/(i+1)}$. 
\textit{Induction step:}
\begin{gather*}
\gamma^{i+1} \overset{\text{I.H.}}{\leq} \gamma 
\frac{1-\frac{i}{i+1}}{1-\gamma\frac{i}{i+1}} \overset{?}{\leq} \frac{1-\frac{i+1}{i+2}}{1-\gamma\frac{i+1}{i+2}} 
\Leftrightarrow \gamma \left(1 - \frac{i}{i+1} - \frac{\gamma}{i+2}\right) \leq 1 - \frac{i+1}{i+2} - 
\frac{\gamma i}{(i+1)(i+2)}  \\
\Leftrightarrow (1-\gamma i)(\gamma - 1) \leq (1-\gamma)^2 \Leftrightarrow -(i-\gamma i) \leq 1- \gamma,
\end{gather*}
where the last inequality holds, because $-(i-\gamma i) \leq 0$ and $(1-\gamma) \geq 0$. I.H. denotes 
\textit{Induction Hypothesis}.
Furthermore,
for $i=1$ 
\begin{align*}
0 < (1-\gamma)^2 = 1- 2 \gamma + \gamma^2  
 \Leftrightarrow \gamma  - \frac{\gamma^2}{2} < \frac{1}{2} \Leftrightarrow \gamma 
< \frac{1-\frac{1}{2}}{1-\frac{\gamma}{2}}
\end{align*}
holds. Hence, the estimator is biased for all $\gamma < 1$. 
It is only unbiased if $\gamma=1$. 

In general, value estimators that fulfill the Bellman equation, respectively use the value function,
must at least be discount dependent to be able to be unbiased for general MRPs, as the following 
Lemma shows: 
\begin{lemma}[p. \pageref{proof:ValueFuncBias}]
\label{lem:ValueFuncBias}
For the MRP from Figure \ref{fig:cyclic_example} (A) and for $n=1$ there exists no parameter estimator $\bar p$
that is independent of $\gamma$ such that $V(\bar p)$ is unbiased for all parameters $p$ and all
discounts $\gamma$.
\end{lemma}

\subsection{Maximum Likelihood Parameter Estimates and Sufficient Statistics}
We start this section with a derivation of the maximum likelihood parameter 
estimates. After that we introduce a minimal sufficient statistics for MRPs and
we show that this statistic equals the maximum likelihood estimates.

\subsubsection{Maximum Likelihood Parameter Estimates}
\label{ML}
Let $p_{ij}$ be the transition probability of state $i$ to $j$,
$p_i$ the probability to start in 
$i$ and $x$ a sample
consisting of $n$ iid state sequences $x_1, \ldots, x_n$. The log-likelihood of the sample is 
\begin{align*}
\log \mathbbm{P}[x|p] = \sum_{k=1}^n \log \mathbbm{P}[x_k|p].
\end{align*}
The corresponding maximization problem is given by
\begin{align*}
\max_{p_{ij},p_i} \sum_{i=1}^n  \log \mathbbm{P}[x_i|p_{ij},p_i] \text{,\quad \quad s.t.:} 
\sum_{j\in \mathbb{S}} p_{ij} = \sum_{j\in \mathbb{S}} p_j = 1.
\end{align*}
The unique solution for $p_{ij}$ and $p_i$  (Lagrange multipliers) is given by
\begin{equation}
\label{MLDef1}
 p_{ij} = \frac{\mu_{ij}}{K_{i}} =: \bar p_{ij} \text{\quad and  \quad} 
 p_{i} = \frac{1}{n} \bigl(K_{i} - \sum_{j \in \mathbb{S}}\mu_{ji}\bigr)
=: \bar p_{i},
\end{equation}
where $K_i$ denotes the number of visits of state $i$, $\mu_{ij}$ the number of direct transitions from $i$ to $j$,
$\bar p_{ij}$ the estimate of the true transition probability $p_{ij}$ and 
$\bar p_{i}$ the estimate of the true starting probability $p_{i}$.

\subsubsection{Sufficient Statistics for the MRP Parameters}
\label{sec:suff_ml}
Information about a sample is typically 
available through a \textit{statistic} $\sS$ of the data (for example $\sS = \sum_i x_i$, where $x$ is a sample). 
A statistic which contains all information about a sample 
is called \textit{sufficient}. Important properties of sufficient statistics are \textit{minimality} and 
\textit{completeness}. The minimal sufficient statistics is the sufficient statistic with
the smallest dimension (typically the same dimension as the parameter space). 
Formally, suppose that a statistic $\sS$ is sufficient for a parameter $\theta$. 
Then $\sS$ is minimally sufficient if $\sS$ is a function of any other statistic $\sT$ that is sufficient for $\theta$.
Formally, a statistic $\sS$ is complete if
$\mathbbm{E}_\theta[h(\sS)] = 0$ for all $\theta$ implies $h = 0$ almost surely.
The theorem from \textit{Rao and Blackwell} \citep{KEND91} states that for a complete and minimal sufficient statistics 
$\sS$ and any unbiased estimator $A$ of a parameter $\theta$ the estimator $\expect[A|\sS]$ is the optimal unbiased 
estimator with respect to any convex loss function and hence the unbiased estimator with minimal MSE.

The maximum likelihood solution is a \textit{sufficient statistics} for the MRP parameters. We demonstrate this 
with the help of the \textit{Fisher-Neyman factorization theorem} \citep{KEND91}. It states that a
statistic is sufficient if and only if the density $f(\mathbf{x}|\mathbf{\theta})$ can be factored into
a product $g(\sS,\mathbf{\theta})h(\mathbf{x})$. 
For a MRP we can factor the density as needed by the Fisher-Neyman theorem ($h(\mathbf{x})=1$ in our case),
\begin{align*}
\mathbbm{P}(\vx|p) 
&= \prod_{i=1}^n \Bigl(p_{\vx_i(1)}  \prod_{j=2}^{L_i} p_{\vx_i(j-1)\vx_i(j)}\Bigr) 
= \prod_{s \in \statespace} p_s^{(K_s - \sum_{s'} \mu_{s's})} \prod_{s,s' \in \statespace} 
p_{ss'}^{K_s \mu_{ss'}}, 
\end{align*}
where $\vx_i(j)$ is the $j$th state in the $i$th path,
$n$ the number of observed paths and $L_i$ the length of  the $i$th path.
$K_s \mu_{ss'}$ is sufficient for $p_{ss'}$ and because sufficiency is sustained by one-to-one mappings \citep{KEND91} 
this holds true also for $\mu_{ss'}$.
The sufficient statistics is \textit{minimal} because the maximum likelihood solution is unique 
\citep{KEND91}\footnote{It is needed to use the minimal parameter set of the MRP to be formally correct.
The minimal sufficient statistics excludes also one value $\mu_{ss'}$, however the missing value is defined
by the other $\mu$'s.}. 
The sufficient statistic is also \textit{complete} because the sample distribution induced by an MRP forms 
an \textit{exponential family} of distributions (Lemma \ref{lem:complete}, page \pageref{lem:complete}). 
A family $\{P_\theta\}$ of distributions is said to form an s-dimensional exponential family if the distributions $P_\theta$ 
have densities of the form
\begin{equation}
p_\theta(x) = \exp\biggl( \sum_{i=1}^s \eta_i(\theta) T_i(x) - A(\theta)\biggr) h(x)
\end{equation}
with respect to some common measure $\mu$ \citep{LEHM98}. Here, the $\eta_i$ and $A$ are real-valued functions 
of the parameters, the $T_i$ are real-valued statistics and $x$ is a point in the sample space. The $\eta$'s are
called \textit{natural parameters}. It is important that the natural parameters are not functionally related. 
In other words no $f$ should exist with $\eta_2  = f(\eta_1)$. If the natural parameters are not functionally related,
then the distribution is complete \citep{LEHM98}. Otherwise, the family forms only a \textit{curved exponential family}
and  a curved exponential family is not complete.

\subsection{Optimal Unbiased Value Estimator}
\label{sec:MVU}
The \textit{Rao-Blackwell theorem} \citep{KEND91} states that for any 
unbiased estimator $A$ the estimator $\expect[A|\sS]$ is the optimal
unbiased estimator \textit{with probability one} (w.p.1), given $\sS$ is a minimal and complete sufficient statistic. 
For the case of value estimation this means that we can use any unbiased value estimator (e.g. the Monte Carlo estimator) 
and condition it with the statistic induced by the maximum likelihood parameter estimate to get the optimal 
unbiased value estimator. 

\begin{theorem}
\label{th:opt_unbiased}
Let $\bar V$ be the first-visit Monte-Carlo estimator and $\sS$ the sufficient and complete statistics for a given MRP. 
The estimator $\expect[\bar V|\sS]$ is unbiased and the optimal unbiased estimator with respect to any convex loss 
function w.p.1.
Especially, it has minimal MSE w.p.1.
\end{theorem}
From now on, we refer to the estimator $\expect[\bar V|\sS]$ as the \textit{Minimum Variance Unbiased estimator (MVU)}. 
For a deterministic reward the estimator $\expect[\bar V|\sS]$ is given by
\begin{gather}
\label{eq:MVU}
\expect[\bar V|\sS] =\frac{1}{|\mathbf{\Pi}(\sS)|} \sum_{\boldsymbol{\pi} \in \mathbf{\Pi}(\sS)} \bar V(\boldsymbol{\pi}),
\end{gather}
where $\boldsymbol{\pi}:=(\pi_1, \ldots, \pi_i)$ denotes a vector of paths, 
$\mathbf{\Pi}(\sS)$ denotes the set of vectors of paths which are consistent 
with the observation $\sS$, $|\cdot|$ is the size of a set and 
$\bar V(\boldsymbol{\pi})$ is the MC estimate for the vector of paths $\boldsymbol{\pi}$. 
Essentially, $\boldsymbol{\pi}$ is an ordered set of paths and it is an element of
$\mathbf{\Pi}(\sS)$ if it produces the observed transitions, starts and rewards. 
The MC estimate is simply the average value for the paths in $\boldsymbol{\pi}$.
The estimator $\expect[\bar V|\sS]$ is thus the average over all paths which could explain 
the (compressed) observed data $\mathcal{S}$. As an example, take the two state MRP from Figure 
\ref{fig:cyclic_example} (A). Assume that an agent starts twice in state $1$, takes 
three times the cycle in the first run and once in the second. The paths which are consistent with
this observation are:
$$ \mathbf{\Pi}(\mathcal{S}) = \{((1,1,1,2),(1,2)), ((1,1,2),(1,1,2)),((1,2),(1,1,1,2)) \}.$$
The MC estimator for the value of a state $s$ does not consider paths which do not hit $s$. 
On the contrary to that the conditioned estimator uses these paths. To see this take a look at 
the MRP from Figure \ref{CounterEx} (A) at p. \pageref{CounterEx}. Assume, that two paths were 
sampled: $(1,2,4)$ and $(2,3)$.
The MC value estimate for state one uses only the first path. Taking a look at 
$$ \mathbf{\Pi}(\mathcal{S}) = \{((1,2,4),(2,3)), (\mathbf{(1,2,3)},(2,4)), ((2,3), (1,2,4)), 
((2,4),\mathbf{(1,2,3)}) \},$$
we see that the conditioned estimator uses the information.

\subsubsection{Costs of Unbiasedness}
The intuition that the MVU uses all paths is, however, not totally correct. 
Let us take a look at the optimal unbiased value estimator of state 
$1$ of the MRP in Figure \ref{fig:cyclic_example} (B)  
for $\gamma = 1$. Furthermore, assume that one run is made and that the path $(2,1,2,3)$ is observed. 
No permutations of this path are possible and the estimate of state $1$ is therefore the MC estimate of 
path $(1,2,3)$, which is $0$. In general, if we make one run and we observe
$i$ transitions from state $2$ to state $1$, then the estimate is $(i-1)$. I.e. we ignore the first transition.
As a consequence, we have on average the following estimate:
$$ (1-p) \sum_{i=1}^\infty (i-1) p^i = p \frac{p}{1-p} = p V_1.$$
The term $p$ is exactly the probability of the event $\{N_1 \geq 1\}$  and the estimator
is conditionally unbiased on this event.
The intuition is, that the estimator needs to ignore the first transition to achieve (conditional) unbiasedness.

Hence, unbiasedness has its price. Another cost beside this loss in information 
is that the Bellman equation cannot be fulfilled. In Section 
\ref{sec:Unbiased_Bellman} we started with Bellman estimators and we showed that
the estimators are biased.  Here, we have a concrete example of an unbiased estimator that does not fulfill 
the Bellman equation, as $\bar V_1 = (i-1) \not = i = \bar V_2$.  For this example this is counterintuitive 
as $p_{12} = 1$ and essentially no difference between the states exists in the undiscounted case.

\subsubsection{Undiscounted MRPs} 
In the undiscounted case permutations of paths do not
change the cumulated reward. For example,
 $\sum_{i=1}^n R_{\pi(i)\pi(i+1)} = \sum_{i=1}^n R_{\pi(\sigma(i))\pi(\sigma(i)+1)}$, 
if $\sigma$ is a permutation of $(1, \ldots, n)$, because the time at which a reward is observed is irrelevant. 
This invariance to permutations implies already a simple fact. We need the following criterion to state this 
fact:
\begin{crit}[Full Information]
\label{crit:full_inf}
A state $s$ has \textit{full information} if, for every successor state $s'$ of $s$ and all paths $\pi$, it holds 
that 
$$\pi(i) = s' \Rightarrow \exists j \text{ with } j<i \text{ and } \pi(j) = s.$$
\end{crit}
$\pi(i)$ denotes the $i$th state in the path.

Let $\boldsymbol{\pi}$ be a vector of paths following the first visit of state $s$ that 
are consistent with the observations. $\bar V(\boldsymbol{\pi})$ is then given
by $(1/|\boldsymbol{\pi}|) \sum_i \sum_j R^{(i)}_{jj+1}$, where $|\boldsymbol{\pi}|$ is the number of
paths contained in $\boldsymbol{\pi}$ and $R^{(i)}_{jj+1}$ is the observed reward in path $i$ at position
$j$. Rearranging the path does not change the sum and the normalizing term. 
Therefore each consistent path results in the same first-visit MC estimate and the 
MVU equals the first-visit MC estimator.
\begin{corollary}
\label{MCIPEqual}
Let $\bar V$ be the first-visit MC estimator and let the value function be undiscounted. 
If the \textit{Full Information Criterion} applies to a state $s$, then
$$\expect[\bar V_s|\mathcal{S}] = \bar V_s.$$
\end{corollary}
The undiscounted setting allows alternative representations of the optimal estimator. As an example,
suppose we observed one path $\pi := (1,1,1,2)$ with reward $R(\pi) = 2 R_{11}+ 1 R_{12}$. 
The optimal estimator is given by $R(\pi)$. 
Alternatively, we can set the reward for a path $\pi$ with $j$-cycles to $R(\pi) := j R_{11} + R_{12}$ and 
define a new probability measure $\hat \prob[\{j \text{ cycles}\}]$ 
such that 
$\sum_{j=0}^\infty j \hat \prob[\{j \text{ cycles}\}]= i$, 
i.e. we average over the set of paths with 0 to ``$\infty$'' many cycles
using the probability measure $\hat \prob[\{j \text{ cycles}\}]$. If this measure is constraint to satisfy
$\sum_{j=0}^\infty j \hat \prob[\{j \text{ cycles}\}]= i$, then
\begin{equation}
\label{eq:undiscounted_repr_mvu}
\sum_{j=0}^\infty \hat \prob[\{j \text{ cycles}\}] (j R_{11} + R_{12}) = i R_{11} + R_{12} = \text{MVU}.
\end{equation}
We pronounce this point here, because the ML value estimator, which we discuss in the next section, can be interpreted in this way.

\subsubsection{Convergence}
Intuitively, the estimator should converge because MC converges in $L^1$ and almost surely. Furthermore, conditioning
reduces norm-induced distances to the true value. This is already enough to follow $L^1$ convergence
but the almost sure convergence is not induced by a norm. We therefore refer to an integral convergence 
theorem which allows us to follow a.s. under the assumption that the MC estimate
is upper bounded by a random variable $Y \in L^1$. Details are given
in Appendix \ref{app:MVU}.
\begin{theorem}[p. \pageref{app:MVU}]
\label{th:MVU_CONV}
$\expect[\bar V|\sS]$ converges on average to the true value. Furthermore, it converges almost surely 
if the MC value estimate is upper bounded by a random variable $Y \in L^1$.
\end{theorem}
Such a $Y$ exists for example, if the reward is upper bounded by $R_{max}$ and if $\gamma < 1$ as in 
this case each MC estimate is smaller than $R_{max} \sum_{i=0}^\infty \gamma^i = R_{max}/(1-\gamma)$.

A MVU algorithm  can be constructed using Equation \ref{eq:MVU}.
However, the algorithm needs to iterate through all possible paths and therefore has an exponential computation time.

\subsection{Least-Squares Temporal Difference Learning}
\label{LSTD_Sec}
In this section we discuss the relation of the MVU to the LSTD estimator. 
The LSTD estimator was introduced by \citep{BRAD96} and extensively analyzed in \citep{BOY98} and \citep{BOY99}. 
Empirical studies showed that LSTD often outperforms massively TD and MC  with respect to
convergence speed per sample size. In this section we support these empirical findings by showing that
the LSTD estimator is equivalent to the MVU for acyclic MRPs and closely related to the MVU for undiscounted MRPs. 
We derive our statements not directly for LSTD, but for the maximum likelihood value  estimator (ML) which is equivalent 
to LSTD (Section \ref{EquLSTD}). The estimator is briefly sketched in \citep{SUTT88}, where it is also shown 
that batch TD(0) is in the limit equivalent to the ML estimator. The estimator is also implicitly used
in the \textit{certainty-equivalence} approach, where a maximum likelihood 
estimate of an MDP is typically used for optimization.

\subsubsection{Maximum Likelihood Estimator}
The ML value estimator is given by $V(\mathbf{\bar P},\mathbf{\bar r})$, where 
$ \mathbf{ \bar P}:= \bigl( \bar p_{ij}\bigr)$ is the maximum likelihood estimate of the transition matrix and
$\mathbf{\bar r}$ is the vector of the maximum likelihood estimates of the expected one step reward.
Hence, the ML value estimator is given by:
\begin{equation}
\mathbf{\bar V} = \sum_{i=0}^\infty \gamma^i \mathbf{\bar P}^i   \mathbf{\bar r}  = 
(\mathbf{I} - \gamma \mathbf{\bar P})^{-1} \mathbf{\bar r},
\label{MLDef2}
\end{equation}
whereas the Moore-Penrose pseudoinverse is used if $\mathbf{\bar P}$ 
is singular (e.g. too few samples).

\subsubsection{Unbiasedness and the MVU}
If an estimator is a function of the sufficient statistic (e.g. $ \bar V=f(\sS)$) then the conditional estimator
is equal to the original estimator, $\bar V = \expect[\bar V|\sS]$. If the estimator $\bar V$ is also unbiased then
it is due to the Rao-Blackwell theorem the optimal unbiased estimator w.p.1. The defined maximum likelihood estimator
is a function of a minimal and complete sufficient statistic. Therefore, the following relation holds
between the ML estimator and the MVU:
\begin{corollary}
\label{Cor:ML_MVU_Unbiased}
The ML estimator is equivalent to the MVU w.p.1, if and only if it is unbiased.
\end{corollary}
The following tow subsections address two cases where ML is unbiased.

\subsubsection{Acyclic MRPs}
\label{LSTD_Opt}
The ML estimator is unbiased in the acyclic case and therefore 
equivalent to the MVU.
\begin{theorem}[p. \pageref{sec:proof_ml}]
\label{theo:LSTD_unbiased}
The ML estimator is unbiased if the MRP is acyclic. 
\end{theorem}
\begin{corollary}
\label{LSTDOpt}
The ML estimator is equivalent to the MVU w.p.1 if the MRP is acyclic.
\end{corollary}

\subsubsection{Undiscounted MRPs}
\label{sec:cyclic_ml}
It is also possible that ML value estimates for specific states are unbiased even if the MRP is
cyclic. One important case in which ML value estimates are unbiased is characterized by 
the Full Information Criterion. If it applies to a state $i$ then 
the normalization $\prob[\{N_i \geq 1 \}]$ does not depend on  the normalizations of the successor states. 
And in a way the problem of Section \ref{sec:norm_prob} does not affect state $i$.

This can be shown by using Theorem 5 from \citep{SING96}, which states that the ML estimator equals 
the first-visit MC estimator if the Full Information Criterion holds and $\gamma=1$.
Furthermore, in this case the first-visit MC estimator is equivalent to the MVU w.p.1 (Corollary \ref{MCIPEqual}).
Hence, ML is unbiased and optimal w.p. 1. We state this as a corollary:
\begin{corollary}
\label{cor:equiv_fullinf}
The ML estimator of a state $i$ is unbiased and equivalent to the MVU w.p.1 if the Full Information Criterion
 applies to state $i$ and if $\gamma=1$.
\end{corollary}
We analyze this effect using a simple MRP and we give two interpretations.
\paragraph{Example: Cyclic MRP - Unbiased}
We start with calculating the bias of ML explicitly for a simple MRP and thus ``verifying'' the Corollary.
The value of state $1$ for the MRP of Figure \ref{fig:cyclic_example} (A)  with modified rewards $R_{11}=1$, 
$R_{12} = 0$
and $\gamma=1$ is
$ V_1 = (1-p) \sum_{i=0}^\infty i p^i. $
The ML estimate for a sample of $n$ paths is 
\begin{equation}
\label{eq:ML_Value_Cycle}
 \bar V_1 = \Bigl(1-\frac{k}{k+n}\Bigr) \sum_{i=0}^\infty  i \Bigl(\frac{k}{k+n}\Bigr)^i = 
\Bigl(1-\frac{k}{k+n}\Bigr) \frac{k/(k+n)}{(1-k/(k+n))^2} = \frac{k}{n}, 
\end{equation}
where $k$ is the number of taken cycles (summed over all observed paths). Therefore 
\begin{align*}
\expect[\bar V_1] = \expect\left[\frac{k}{n}\right] = \frac{1}{n} \sum_{i=1}^n \expect[k_i].
\end{align*}
Furthermore,
$$ \expect[k_i] = (1-p) \sum_{k_i=0}^\infty  k_i p^{k_i} = V_1$$ 
and the ML estimator is unbiased. Now, Corollary \ref{Cor:ML_MVU_Unbiased} tells us that the ML estimator is equivalent to the MVU w.p.1.

It is also possible to show this equivalence using simple combinatorial arguments.
The MVU and the MC estimate for this MRP is $\frac{k}{n}$:  Let
$u$ be the number of ways how $k$ can be split onto $n$-paths. For each split the summed reward is 
$k$ and the MC estimate is therefore $\frac{k}{n}$. Hence, the MVU is $\frac{u k/n}{u} = \frac{k}{n}$.

\paragraph{Interpretation I: Non-linearity vs. Underestimated Parameters}
It is interesting that ML is unbiased in this example.
In general nonlinear transformations of unbiased parameter estimates produce biased estimators, as
$$\expect[f(\bar \theta)] = f(\theta) = f(\expect[ \bar \theta]) $$
essentially means that $f$ is a linear transformation as $f$ and $\expect$ commute. Furthermore,
the value function is a nonlinear function. Yet, in our example the parameter estimator $\bar \theta$ 
is actually not unbiased. For $n=1$: 
\begin{align*} 
\expect\left[\frac{k}{k+1}\right] = (1-p) \sum_{k=0}^\infty \frac{k}{k+1} p^k 
< (1-p) \sum_{k=1}^\infty p^k 
= (1-p) \sum_{k=0}^\infty p^{k+1} = p. 
\end{align*}
The parameter is underestimated on average. The reason for this lies in the dependency between the visits 
of state 1. For a fixed number of visits, respectively for \textit{iid} 
observations the parameter estimate would be unbiased. 
The relation between these two estimation settings  is very similar to the first-visit and every-visit MC setting.
The first-visit MC estimator is unbiased because it uses only one observation per path
while the every-visit MC estimator is biased.
In our case, the effect is particularly paradox as for the \textit{iid} case the value estimator is biased.

\paragraph{Interpretation II: Consistency of the Set of Paths}
\label{sec:RelMLMVU}
The ML estimator differs in general from the MVU because it uses paths that are inconsistent with 
the observation $\sS$.
For example, given the MRP from Figure \ref{fig:cyclic_example} (A) with modified rewards $R_{11} = 1$, $R_{12} = 0$ and  the observation 
$(1,1,1,2)$. The set of paths consistent with this observation is again 
$\{(1,1,1,2)\}$. The ML estimator, however, uses the following set of paths
$$  \{(1,2),(1,1,2),(1,1,1,2),(1,1,1,1,2) \ldots \},$$
with a specific weighting $\hat \prob[\{j \text{ cycles}\}]$ for a path that contains $j$ cycles. In general, this representation
will result in an estimate that is different from the MVU estimate. However, if Corollary \ref{cor:equiv_fullinf} applies
then both representations are equivalent. The ML estimator can under the assumptions of the corollary be 
represented as a sum over the cycle times with each summand being a product between the estimated path probability and the reward 
of the path. One can see this easily for the example (one run with $i=2$ cycles being taken): 
The path probability is in this case simply $ \hat \prob[\{j \text{ cycles} \}] = \bar p^j (1-\bar p)$
and because $\sum_{j=0}^\infty j \bar p^j (1-\bar p) = i = 2$  (Eq. \ref{eq:ML_Value_Cycle} with $n=1$) 
the estimate is equal to $2 R_{11} + R_{12}$ which is exactly the MVU estimate (compare to eq. \ref{eq:undiscounted_repr_mvu}
on p. \pageref{eq:undiscounted_repr_mvu}).

\subsubsection{Which Estimator is better? The MVU or ML?}
The MVU is optimal in the class of unbiased estimators. However, this does not mean that the ML 
estimator is worse than the MVU. The ML estimator is also a function of the sufficient statistics, 
it is just not unbiased. To demonstrate this, we present two examples based on the MRP from 
Figure \ref{fig:cyclic_example} (A) in Appendix \ref{sec:counter_mvu_ml} (p. \pageref{sec:counter_mvu_ml}). One for 
which the MVU is superior and one where the ML estimator is superior. We summarize this in a corollary:
\begin{corollary}
MRPs exist in which the MVU has a smaller MSE than the ML estimator and MRPs exist
in which the ML estimator has a smaller MSE than the MVU.
\end{corollary}

\subsubsection{The LSTD Estimator}
The LSTD algorithm computes analytically the parameters which minimize the empirical quadratic error for 
the case of a linear system. 
\citep{BRAD96} showed that the resulting algorithm converges almost surely to the true value. 
In \citep{BOY98} a further characterization of the least-squares solution is given.
This turns out to be useful to establish the relation to the ML value estimator. 
According to this characterization, the LSTD estimate $\bar V$ 
is the unique solution of the Bellman equation, i.e. 
\begin{equation}
\label{LSTDCons}
\mathbf{\bar V} = \mathbf{\bar r} + \gamma \mathbf{\bar P} \mathbf{\bar V},
\end{equation}
where $\mathbf{\bar r}$ is the sample mean estimate of the reward and $\mathbf{\bar P}$ is the maximum likelihood 
estimate of the transition matrix.

\label{EquLSTD}
Comparing Equation \ref{LSTDCons} with Equation \ref{MLDef2} of the ML estimator it
becomes obvious that both are equivalent if 
the sample mean estimate of the reward equals the maximum 
likelihood estimate. 
\begin{corollary} 
The ML value estimator is equivalent to LSTD if the sample mean and the maximum likelihood estimator of the expected reward
are equivalent.
\end{corollary}

\subsection{Monte Carlo Estimation}
\label{sec:assumpt-free-estim}
\label{MCRelSec}
We first summarize Theorem 5 from \citep{SING96}  
and Cor. \ref{MCIPEqual} from p. \pageref{MCIPEqual}:
\begin{corollary}
\label{cor:MC_Opt}
The (first-visit) MC estimator  of a state $i$ is equivalent to the MVU and to the ML 
estimator w.p.1 if the Full Information Criterion applies to state $i$ and an undiscounted MRP is given.
\end{corollary}
Essentially, the corollary tells us that in the undiscounted case it is only the ``amount'' of information 
that makes the difference between the MC estimator and the MVU, respectively the ML
estimator. Amount of information refers here to the observed
paths. If MC observes every path then the estimators are equivalent.

From a different point of view this tells us that in the undiscounted case the MRP structure is only useful
for passing information between states, but yields no advantage beyond that.

\subsubsection{Discounted MRPs}
\label{sec:Disc_MC_counter}
In the discounted cyclic case the MC estimator differs from the ML and the MVU estimator. It differs from ML because 
ML is biased. The MC estimator is equivalent to the MVU in the undiscounted case because the order in which the
reward is presented is irrelevant. That means the time at which a cycle occurs is irrelevant. In the
discounted case this is not true anymore. Consider again the MRP from Figure \ref{fig:cyclic_example} (A) with
rewards $R_{11} =1$, $R_{12} =0$ and 
the following two paths $\boldsymbol{\pi} = ((1,1,1,2),(1,2))$. The MC estimate is $1/2((1 + \gamma) + 0)$.
The set of paths consistent with this observation is $\boldsymbol{\Pi}(\sS) = 
\{((1,1,1,2),(1,2)), ((1,1,2),(1,1,2)),
((1,2),(1,1,1,2))\}$.  Hence, the MVU uses the cycle $(1,1,2)$ besides the observed ones.
The MVU estimate is  $ 1/3 ( (1+\gamma)/2 + 2/2  + (1+\gamma)/2) = 1/3(2+\gamma)$.
Both terms are equivalent if and only if $\gamma=1$. For this example the Full Information
Criterion applies. 

Similarly, for acyclic MRPs the MC estimator is different from the ML/MVU  estimator if $\gamma < 1$. 
Consider a 5 state MRP 
with the following observed paths: $((1,3,4),(1,2,3,5))$, a reward of $+1$ for $3\rightarrow 4$
and $-1$ for $3\rightarrow 5$. The ML estimate is $(1/4 \gamma^2 + 1/4 \gamma)(1 -1) = 0$, while 
the MC estimate is $1/2 ( - \gamma^2  + \gamma)$ which is $0$ if and only if $\gamma=1$.
Again the Full Information Criterion
applies. 

\subsubsection{Ordering with Respect to other Value Estimators}
Beside the stated equivalence the MVU is for every MRP at least as good as the first-visit MC estimator, because
the first-visit MC estimator is unbiased. The relation to ML is not that clear cut. In general MRPs exist 
where the first visit MC estimator is superior and MRPs exist where the ML estimator is superior 
(See Appendix \ref{sec:counter_mvu_ml}, p. \pageref{sec:counter_mvu_ml} for examples). How about TD($\lambda$)? Again
the relation is not clear cut. In the case that the MRP is acyclic and that Corollary 
\ref{cor:MC_Opt} applies the first-visit MC estimator is at least as good as TD($\lambda$). In general,
however, no ordering exists (See Appendix \ref{Counter:MC_TD}, p. \pageref{Counter:MC_TD} for examples).

\subsection{Temporal Difference Learning}
\label{sec:appr-equal-weight}
One would like to establish inequalities between the estimation error of TD and the error of other estimators
like the MVU or the ML estimator. For the acyclic case TD($\lambda$) is essentially unbiased and the MVU and the ML 
estimator are superior to TD.
However, for the cyclic case the analysis is not straightforward, as TD($\lambda$) is biased for $\lambda < 1$ 
and does not fulfill the Bellman equation. So TD is in a sense neither in the estimator class of the MVU nor of the ML 
estimator and conditioning with a sufficient statistics does not project TD to either of these estimators. 

The bias of TD can be verified with the MRP from Figure \ref{fig:cyclic_example} (A) with rewards
$R_{11} = 1$, $R_{12} = 0$, with a discount  of $\gamma=1$  and with $n=1$. 
If we take the TD(0) estimator with  a learning rate of $\alpha_j= 1/j$ then
the value estimate for state $0$ is $i/(i+1) \sum_{j=1}^i 1/j$ if $i$ cyclic transitions have been observed. 
The estimate should on average equal $i$ to be unbiased. Yet, for $i>0$ it is strictly smaller than $i$.

While our tools are not usable to establish inferiority of TD, we can still interpret the weaknesses of TD with it.
In the following we focus on the TD(0) update rule.

\subsubsection{Weighting of Examples and Conditioning}
In the examples comparing TD($\lambda$) and MC (Section \ref{CounterTD} p. \pageref{CounterTD})
one observes that a weakness of TD(0) is that not all of the examples are weighted 
equally. In particular, Equation \ref{eq:lin_est} on page \pageref{eq:lin_est} suggests that no 
observation should be preferred over another. Intuitively, conditioning suggests so too: 
For an acyclic MRP  TD(0) can be written as $\bar V_i = \tilde p_{ij} (R_{ij} + \gamma \bar V_j)$,
whereas $\tilde p_{ij}$ differs 
from the maximum likelihood parameter estimates $\bar p_{ij}$
due to the weighting. 
Generally, conditioning with a sufficient statistics permutes the order of the observations and resolves
the weighting problem. Therefore, one would assume that conditioning with the element $\bar p_{ij}$ of the 
sufficient statistics changes $\bar V_i$ to $\bar p_{ij} (R_{ij} + \gamma \bar V_j)$. As conditioning improves
the estimate, the new estimator would be superior to TD(0). However, conditioning with just a single 
element $\bar p_{ij}$ must not modify the estimator at all, as the original path might be reconstructed from
the other observations. E.g. if one observes a transition  $1\rightarrow 2$  and $2 \rightarrow 3$,
with $2 \rightarrow 3$ being the only path from state 2 to state 3, then it is enough to know that 
transition $1\rightarrow 2$ occurred and state $3$ was visited.

Despite these technical problems, the superiority of $\bar p_{ij}$ over $\tilde p_{ij}$ and the weighting 
problem are reflected in the contraction properties of TD(0). Due to \citep{SUTT88} TD(0) contracts towards 
the ML solution. Yet, the contraction is slow compared to the case where each example is weighted equally.

\subsubsection{Weighting of Examples and Contraction Factor}
We continue with  another look at the familiar ML  equation:
$  
\mathbf{\bar V} = \mathbf{\bar r} + \gamma \mathbf{\bar P} \mathbf{\bar V} =: \mathbf{\bar T} \mathbf{\bar V}.
$
If the matrix $\eP$ is of full rank then the ML estimate is the sole fixed point of the Bellman operator 
$\mathbf{\bar T}$. The ML estimate can be gained by solving the equation, i.e 
$\mathbf{\bar V} = (\mathbf{I} - \gamma \eP)^{-1} \mathbf{\bar r}$. 
Alternatively, it is possible to make a fixed point iteration. I.e. starting with an initial 
guess $\mathbf{\bar V}^{(0)}$ and iterating the equation, i.e $\mathbf{\bar V}^{(n)} = \mat{\bar T} 
\mathbf{\bar V}^{(n-1)}$. Convergence to the ML solution is guaranteed by the 
\textit{Banach Fixed Point Theorem}, because $\mat{\bar T}$ is a contraction. The contraction
factor is upper bounded by $\gamma || \eP|| \leq \gamma $, where $||\cdot||$ denotes in the 
following the \textit{operator norm}. The bound can be improved by using better suited norms (e.g. 
\citep{BERT96}). Hence, for $n$ updates the distance to the ML solution is reduced by a factor 
of at least $\gamma^n$. 

\label{RelTD}
Applying the TD(0) update (Eq. \ref{TD0Def}) to the complete value estimate $\mathbf{\bar V}$ using $\mathbf{\bar P}$ and a 
learning rate of $1/n$ results in
$$ \mathbf{\bar V}^{(n)} =   \mathbf{\bar V}^{(n-1)} + \frac{1}{n}\left( \mathbf{r} + \gamma \eP 
 \mathbf{\bar V}^{(n-1)} -  \mathbf{\bar V}^{(n-1)}\right)=
\left(\frac{n-1}{n}  + \frac{1}{n} \mathbf{\bar T}\right) 
\mathbf{\bar V}^{(n-1)}. $$ 
In this equation the weighting problem becomes apparent: The contraction $\mathbf{\bar T}$ affects
only a part of the estimate. Yet, the operators $\mathbf{\bar S}^{(n)}:=
\left(\frac{n-1}{n}  + \frac{1}{n} \mathbf{\bar T}\right)$ are still contractions. For  
$\mathbf{\bar V}$ and  $\mathbf{\bar W}$:  
$$ ||\mathbf{\bar S}^{(n)} \mathbf{\bar V} -  \mathbf{\bar S}^{(n)} \mathbf{\bar W}||  \leq
\frac{n-1}{n} || \mathbf{\bar V} -  \mathbf{\bar W} ||  + 
\frac{1}{n} || \mathbf{\bar T}|| || \mathbf{\bar V} -  \mathbf{\bar W} ||  
\leq  \frac{n-1 + \gamma}{n}    || \mathbf{\bar V} -  \mathbf{\bar W} ||.  $$
The contraction coefficient is therefore at least $\frac{n-1 + \gamma}{n}$. The ML solution (in the following 
$\mathbf{\bar V}$)
is a fixed point for the $\mathbf{\bar S}^{(i)}$ and for $n$ iterations the distance is bounded by
$$ || \mathbf{\bar S}^{(n)} \ldots \mathbf{\bar S}^{(1)} \mathbf{\bar V}^{(0)}-  \mathbf{\bar V}||
\leq
\frac{\prod_{i=0}^{n-1}(i+\gamma)}{n!} || \mathbf{\bar V}^{(0)} - \mathbf{\bar V}||.
$$
The smaller $\gamma$ the faster the contraction. Yet, even in the limit the contraction 
is much slower than the contraction with the ML fixed point iteration, i.e. for $\gamma =0$
the distance decreases at least with $1/n$ while for the ML fixed point iteration it decreases
with $\gamma^n$. For $\gamma= 0.1$  and two applications of the Bellman operator
the contraction is at least $\gamma^2 = 1/100$ and it needs 100 iterations with the TD(0) equation
to reach the same distance.

TD(0) is applied only to the current state and not to the full value vector. The same can be done 
with the ML fixed point iteration, i.e. $\bar V_i = \bar p_{ij}( \bar R_{ij} + \gamma \bar V_j)$. 
We analyze the contraction properties of this estimator in the empirical part and 
we refer to the estimator as the iterative Maximum Likelihood (iML) estimator. 
The costs of the algorithm are slightly higher than the TD(0) costs: O$(|\statespace|)$ (time) and O$(|\statespace|^2)$
(space).

The restriction to the current path does not affect the convergence, i.e. the restricted iteration converges 
to the ML solution. Intuitively, the convergence is still guarantied, as a contraction of $\gamma$ is achieved by visiting 
each state once and because each state is visited infinitely often. Using that idea the following Theorem can
be proved:
\begin{theorem}
\label{th:AETD_Unbiased_Conv}
iML is unbiased for acyclic MRPs, converges on average and almost surely to the true value.
\end{theorem}
We use this algorithm only for the analysis  and we therefore omit the proof.

\subsection{Summary of Theory Results}
\label{sec:summary_theory}
We conclude the theory section with two tables that summarize central properties of estimators and established 
orderings. Footnotes are used to reference the corresponding theorems, corollaries or sections.
We start with a table that summarizes the properties of the different estimators (Table \ref{tab:sum_1}). The row \textbf{Optimal}
refers to the class of unbiased estimators and to convex loss functions. The statement that ML is unbiased if  
the Full Information Criterion
is fulfilled and $\gamma=1$ applies state wise. I.e. for a cyclic MRP there will   
exist a state for which the ML estimator is biased. However, if 
the Full Information Criterion applies to a state, then 
the ML estimator for this particular state is unbiased. Finally, F-visit MC denotes the first-visit Monte-Carlo estimator.

\begin{table}[h!]
\begin{tabular}{|p{1.7cm}|p{2.4cm}|p{2.4cm}|p{1.2cm}|p{2.35cm}|}
\hline
Estimator & \textbf{MVU} & \textbf{ML/LSTD} &   \textbf{TD($\lambda$)} & \textbf{(F-visit) MC} \\ 
\hline
Convergence & \lp{1}, a.s.$^{(1)}$  & \lp{1}, a.s.  &   \lp{1}, a.s.
& \lp{1}, a.s.  \\ 
\hline
Cost (Time) & $\exp$?$^{(2)}$ & O$(|\statespace|^3)$ &    O$(|\statespace|)$ &  O$(|\statespace|)$ \\ 
Cost (Space)& & O$(|\statespace|^3)$ &    O$(|\statespace|)$ & O$(|\statespace|)$ \\  
\hline
 Unbiased & $\surd^{(3)}$  &  Acyclic$^{(4)}$ or  Cr. \ref{crit:full_inf}  and $\gamma=1^{(5)}$ 
& Acyclic$^{(6)}$
& $\surd$  \\
\hline 
 Bellman & Acyclic$^{(4)}$ or Cr.  \ref{crit:full_inf} and $\gamma=1^{(5)}$  &  $\surd$ & & \\
\hline 
Optimal  & $\surd^{(3)}$  & Acyclic$^{(4)}$ or  Cr. \ref{crit:full_inf}  and $\gamma=1^{(5)}$ 
&  &  
Cr. \ref{crit:full_inf}  and $\gamma=1$$^{(7)}$ \\
\hline
\end{tabular}
\caption{Comments and references: (1) Th. \ref{th:MVU_CONV}, p. \pageref{th:MVU_CONV}. 
(2) Eq. \ref{eq:MVU}, p. \pageref{eq:MVU}.
(3) Th. \ref{th:opt_unbiased}, p. \pageref{th:opt_unbiased}.
(4) Cor. \ref{LSTDOpt}, p. \pageref{LSTDOpt}.
(5) Cor. \ref{cor:equiv_fullinf}, p. \pageref{cor:equiv_fullinf}.
(6) Minorly modified TD estimator. Th. \ref{th:TDlam_unbiased}, p. \pageref{th:TDlam_unbiased}.
(7) Cor. \ref{cor:MC_Opt}, p. \pageref{cor:MC_Opt}. 
Counterexamples for $\gamma<1$: Sec. \ref{sec:Disc_MC_counter}  
p.\pageref{sec:Disc_MC_counter}.
}
\label{tab:sum_1}
\end{table}

Table \ref{tab:sum_2} summarizes established orderings between value estimators. The legend is the following: 
$=$ means equivalent, $\not =$ means not comparable, $\leq$ means that the estimator in the corresponding row has a 
smaller risk (estimation error) than the estimator in the corresponding column. With \textbf{In general} we mean for 
$\not =$ that there exist MRPs where the row estimator is superior and MRPs where the column estimator is superior. However,
for a subclass of MRPs, like acyclic MRPs, one of the estimators might be superior or they might be equivalent. For $\leq$ 
\textbf{in general} means that the row estimator is always as good as the column estimator, however, both might be equivalent 
on a subclass of MRPs.
\begin{table}[h!]
\begin{center}
\begin{tabular}{|p{1.8cm}|p{2.8cm}|p{2.5cm}|p{3.3cm}|}
\hline
 & \textbf{ML/LSTD} &   \textbf{TD($\lambda$)} & \textbf{(F-visit) MC} \\ 
\hline
\textbf{MVU} & 
\begin{minipage}[t]{\linewidth}
ML unbiased: $\bm{=}^{(1)}$\\
In general: $\bm{\not =}^{(2)}$  
\end{minipage}
 & Acyclic: $\bm{\leq}^{(3,4)}$ & 
\begin{minipage}[t]{\linewidth}
Cr. \ref{crit:full_inf} and $\gamma=1$: $\bm =^{(5)}$ \\
In general: $\bm \leq^{(4)}$
\end{minipage}
 \\
\hline
\textbf{ML/LSTD} & \cellcolor[gray]{0.3} &   Acyclic: $\bm{\leq}^{(3,4,7)}$  &
\begin{minipage}[t]{\linewidth}
Cr.  \ref{crit:full_inf} and $\gamma=1$: $\bm =^{(6)}$ \\
In general: $\bm{\not =}^{(2)}$
\end{minipage} \\
\hline
\textbf{TD($\lambda$)}  & 
\cellcolor[gray]{0.3}
 &  \cellcolor[gray]{0.3} & In general: $\bm{\not =}^{(8)}$   \\
\hline
\end{tabular}
\end{center}
\caption{Comments and references: (1) Cor. \ref{Cor:ML_MVU_Unbiased}, p. \pageref{Cor:ML_MVU_Unbiased}.
(2) Counterexamples: App. \ref{sec:counter_mvu_ml}, p. \pageref{sec:counter_mvu_ml}.
(3) Minorly modified TD estimator. Th. \ref{th:TDlam_unbiased}, p. \pageref{th:TDlam_unbiased}.
(4) Th. \ref{th:opt_unbiased}, p. \pageref{th:opt_unbiased}.
(5) Cor. \ref{cor:MC_Opt}, p. \pageref{cor:MC_Opt}. 
(6) Th. 5 in \citep{SING96}.
(7) Cor. \ref{LSTDOpt}, p. \pageref{LSTDOpt}.
(8) Counterexamples: App. \ref{Counter:MC_TD}, p. \pageref{Counter:MC_TD}.
}
\label{tab:sum_2}
\end{table}

\section{Comparison of Estimators: Experiments}
\label{sec:emp_estimator-comparison}
In this section we make an empirical comparison of the estimators. We start with a comparison using acyclic MRPs. 
For this case the ML estimator equals the MVU and the MVU solution can be computed efficiently. This allows
us to make a reasonable comparison of the MVU/ML estimator with other estimators. In a second set of
experiments we compare the MVU with the ML estimator using a very simple cyclic MRP. 
In a final set of experiments we compare the contraction properties of iML and TD(0).

\subsection{Acyclic MRPs}
\label{ExpEval}
We performed three experiments for analyzing the estimators. 
In the first experiment we measured the MSE in dependence to the number of observed paths. 
In the second experiment we analyzed how the MRP structure affects the estimation performance.
As we can see from Corollary \ref{MCIPEqual} the performance difference 
between ``MDP''  based estimators such as TD or ML and model free estimators like MC 
 depends on the ratio between the number of  sequences hitting a state $s$ itself 
and the number of sequences entering the subgraph of successor states without hitting $s$. 
We varied this ratio in the second experiment and measured the MSE. 
The third experiment was constructed to analyze the practical usefulness of the different estimators. We measured
the MSE in relation to the calculation time.

\paragraph{Basic Experimental Setup}
We generated randomly acyclic MRPs for the experiments.
The generation process was the following:
We started by defining a state $s$ for which we want to estimate the value. Then we 
generated randomly a graph of successor states. We used different layers with a random number of states in each layer. 
Connections
were only allowed between adjacent layers. Given these constraints, the transition matrix was generated randomly 
(uniform distribution). For the different experiments, a specific number of starts in state $s$ was defined.
Beside that, a number of starts in other states were defined. Starting states were all states in the first layers 
(typically the first 4). Other layers which were further apart from $s$ were omitted as paths starting in these
contribute few to the estimate, but consume computation time. The distribution over the starting states was chosen to 
be uniform.
Finally, we randomly defined rewards for the different transitions (between 0 and 1), while a 
small percentage  (1 to 5 percent) got a high reward (reward 1000). Beside the reward definition, 
this class of MRPs contains a wide range of acyclic MRPs.    
We tested the performance (empirical MSE) of the ML, iML, MC and TD estimators.
For the first two experiments the simulations were repeated 300 000 times for each parameter setting. We splitted
these runs into 30 blocks with 10 000 examples each and calculated the mean and standard deviation for these.
In the third experiment we only calculated the mean using 10 000 examples.
We used the modified TD(0) version which is unbiased with a learning rate of $1/i$ for each state. The 
ML solution was computed at the end and not at each run. This means no intermediate estimates were available, which 
can be a drawback. We also calculated the standard TD(0) estimates. The difference to the modified TD(0) version 
is marginal and therefore we did not include the results in the plots.

\begin{figure}[h!]
\begin{center}
\setlength{\epsfxsize}{3.in}
\centerline{\epsfbox{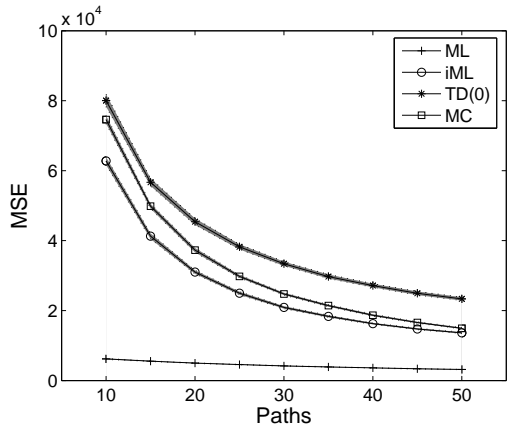}}
\caption{MSE of ML, iML, TD(0) and MC in relation to the number of observed paths.  
 The state space consisted of 10 layers with 20 states per layer.}
\label{Sim:MSE_N}
\end{center}
\end{figure} 

\subsubsection{Experiment 1: MSE in Relation to the Number of Observed Paths}
In the first experiment, we analyzed the effect of the number of observed paths given a fixed rate of $p_s =0.2$
for starts in state $s$. The starting probability for state $s$ is high and beneficial to MC (The effect of 
$p_s$ is analyzed in the second experiment). Apart from ML, all three estimators perform quite similarly with a small
advantage for iML and MC (Figure \ref{Sim:MSE_N}). 
ML is even for few paths strongly superior and the estimate is already good for 10 paths. 
Note that, due to the scale the improvement of ML is hard to observe.

\subsubsection{Experiment 2: MSE in Relation to the Starting Probability}
\begin{figure}[h!]
\begin{center}
\setlength{\epsfxsize}{3in}
\centerline{ \epsfbox{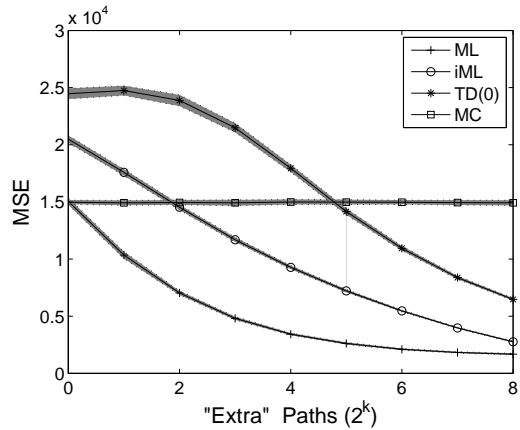}}
\caption{MSE of ML, iML, TD(0)  and MC in relation to the starting probability of the estimated state. 
The state space consisted of 10 layers with 20 states per layer. }
\label{Sim:StartProb}
\end{center}
\end{figure} 
In the second experiment we tested how strongly the different estimators use the Markov structure. 
To do so, we varied the ratio of starts in state $s$ (the estimator state) to starts in the subgraph. 
The paths which start in the subgraph can only improve the estimation quality of state $s$ if
the Markov structure is used.
Figure \ref{Sim:StartProb} shows the results of the simulations. 
The $x$-axis gives the number of starts in the subgraph while the number of starts
in state $s$ was set to $10$. We increased the number exponentially.
The exponential factor is printed on the x-axis.
$x=0$ is equivalent to always start in $s$.
One can see that the MC and ML estimator are equivalent if in each run the path
starts in $s$. Furthermore, for this case MC outperforms TD due to the weighting problem of TD (Section \ref{RelTD}). 
Finally, TD, iML  and ML make a strong use of paths which does not visit state $s$ itself.
Therefore, TD becomes superior to MC for a higher number of paths. The initial plateau for the TD estimator appeared
in the modified and the standard version.  We assume that it is an effect of the one step error propagation 
of TD($0$). For the one step error propagation a path starting in a state $s'$ in the $ith$ layer
can only improve the estimate if $i$ paths are observed that span the gap between $s$ and $s'$. 
The probability of such an event is initially very small but increases with more  paths.

\subsubsection{Experiment 3: MSE in Relation to Calculation Time}
\begin{figure}[h!]
\begin{center}
\setlength{\epsfxsize}{5.5in}
\centerline{\epsfbox{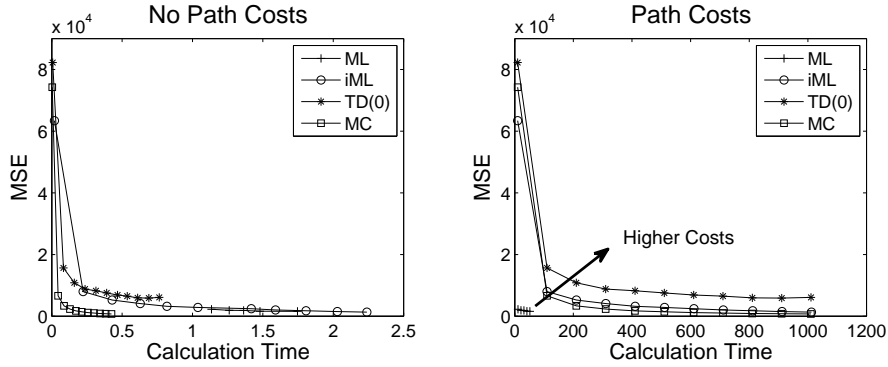}}
\caption{MSE in relation to the computation time of the ML, iML, TD(0) and MC estimator. 
The left plot shows pure computation time (we excluded computation time for MRP calculations like state changes).
In the right plot, an extra factor for each observed path is included (one second per path).
The state space consisted of 10 layers with 20 states per layer. We tracked for a given 
number of paths (ML: 10-50, iML, TD(0), MC: 10-1000) the MSE and the computation time. The plot was constructed 
with the mean values for every number of paths.}
\label{Sim:Numeric}
\end{center}
\end{figure} 
\label{exp:comp_time}
In many practical cases the convergence speed per sample is not the important measure. It is   
the convergence speed per time that is important. The time needed for reaching a specific MSE level consists of the MSE for 
a given number of paths, the costs to calculate the estimate from the sample, and the costs for generating 
the paths. We constructed an experiment to evaluate this relation (Figure \ref{Sim:Numeric}). We first tested
which estimator is superior if only the pure estimator computation time is regarded (left part). 
For this specific MRP the MC estimator converges fastest in dependence of time.
The rate for starts in  state $s$ was 0.2, which is an advantage for MC. The ratio will typically be much lower.
The other three estimators seem to be more or less equivalent. In the second plot a constant cost of 1 was introduced 
for each path. Through this the pure computation time becomes less important while the needed number of paths
for reaching a specific MSE level becomes relevant. As ML needs only very few paths, it becomes superior 
to the other estimators. Further, iML catches up on MC. For higher costs the estimators
will be drawn further apart from ML (indicated by the arrow). 
The simulations suggest that MC or TD (dependent on the MRP) are a 
good choice if the path costs are low. For higher costs ML and iML are alternatives.

\subsection{Cyclic MRPs: MVU - ML Comparison}
\begin{figure}[h!]
\begin{center}
\setlength{\epsfxsize}{5.5in}
\centerline{\epsfbox{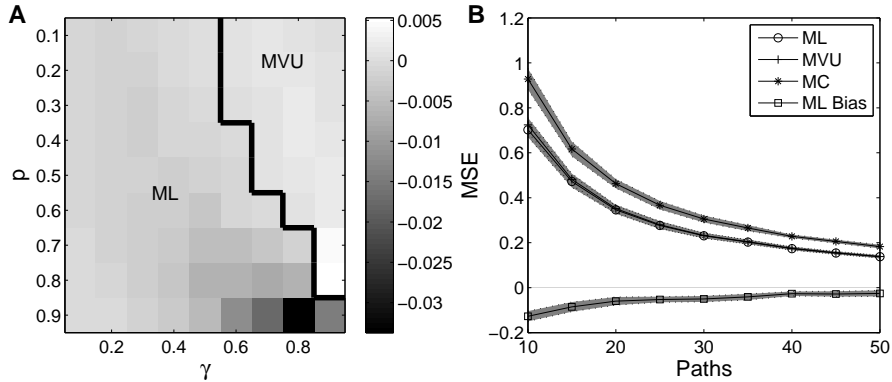}}
\caption{
\textbf{A}: The plot shows the difference in MSE between the ML estimator and the MVU (MSE(ML)- MSE(MVU)) 
for 10 paths and different values of $\gamma$ and $p$. In the top right part the MVU is superior and in the 
remaining part the ML estimator. \textbf{B}: The plot shows the MSE of the ML, the MVU and the MC estimator
and the bias of ML in dependence of the number of paths for $p=\gamma = 0.9$. 30 000 samples were used for
the mean and the standard deviation (30 blocks with 1000 examples).
\label{fig:cyclic_mvu_ml}
}
\end{center}
\end{figure} 
\label{exp:cyclic}
Calculating the MVU is infeasible without some algebraic rearrangements. Yet, the algebraic rearrangements
get tricky, even for simple MRPs. We therefore restrict the comparison of the MVU and the ML estimator 
to the simplest possible cyclic MRP, i.e. the MRP from Figure \ref{fig:cyclic_example} (A) with
rewards $R_{11} = 1$ and $R_{12} = 0$.
The MC and ML value estimates are  
$$\frac{1}{n} \sum_{u=1}^n \sum_{j=0}^{i_u} 
\gamma^{i_u} = \frac{1}{1-\gamma} - \frac{1}{n} \sum_{u=1}^n \frac{\gamma^{i_u+1}}{1-\gamma}$$
and
$$ (1- \bar p) \sum_{i=0}^\infty \frac{1-\gamma^{i+1}}{1-\gamma} 
\bar p^i = \frac{1}{1-\gamma} - \frac{\gamma}{1-\gamma} \frac{1-\bar p}{1-\gamma \bar p}, $$
where $i_u$ denotes the number of times the cycle has been taken in run $u$.
The MVU sums the MC estimates over all consistent sets, i.e. over all vectors $(k_1, \ldots, k_n)$
which fulfill $\sum_{u=1}^n k_u = s := \sum_{u=1}^n i_u$. Let the normalization $\mathcal{N}$ 
being the size of this set and $MC(k_i)$ being the MC estimate for $k_i$ cycles. The MVU is given by
\begin{align*}
&\frac{1}{n\mathcal{N}} \sum_{(k_1, \ldots, k_n) = s} MC(k_1) + \ldots  +MC(k_n) \\
&=   \frac{1}{n\mathcal{N}} \sum_{k_1=0}^s \ldots \sum_{k_{n-1}=0}^{s - k_1 \ldots - k_{n-2}}
MC(k_1) + \ldots + MC(k_n), 
\end{align*}
where in the second line $k_n = s - k_1 \ldots - k_{n-1}$.
The number of times $k_u$ takes a value $j$ is independent of $u$, i.e. MC($k_1$) appears equally often 
as MC($k_i$) if $k_1 = k_i$. Hence, it is enough to consider $MC(k_1)$  and the MVU is
$$  \frac{1}{\mathcal{N}} \sum_{k_1=0}^s MC(k_1) 
\sum_{k_2=0}^{s-k_1} \ldots \sum_{k_{n-1}=0}^{s - k_1 \ldots - k_{n-2}} 1 =:  
\frac{1}{\mathcal{N}} \sum_{k_1=0}^s MC(k_1) \mathcal{C}(k_1). $$
Finally, the coefficient is $\mathcal{C}(k_1) = {s + n -2 - k_1 \choose n -2 }$ 
and the normalization is $\mathcal{N} = {s + n -1  \choose n -1 }$. 
The derivation can be done in the following way. First, observe that  $1 = {k_n  \choose 0 }$. Then,
that $\sum_{k_{n-1}=0}^{s - k_1 \ldots - k_{n-2}} {k_n \choose 0} = {1+ (s - k_1 \ldots - k_{n-2}) \choose 1}$
(e.g. rule 9 in \citep{AIGNER06} p. 13). And finally that 
$\sum_{k_{n-2}=0}^{s - k_1 \ldots - k_{n-3}} {1+ (s - k_1 \ldots - k_{n-2}) \choose 1}
= \sum_{k_{n-2}=0}^{s - k_1 \ldots - k_{n-3}} {1+  k_{n-2} \choose 1}$. Iterating the steps leads to the normalization
and the coefficients. In summary the MVU is 
\begin{equation}
\label{eq:MVU_Simple}
 \frac{1}{1-\gamma} - \frac{1}{(1-\gamma) {s+n-1\choose n-1}}  \sum_{i=0}^s {s + n-2 - i \choose n-2}\gamma^i. 
\end{equation}
We compared the MVU to the ML estimator in two experiments. The results are shown in Figure \ref{fig:cyclic_mvu_ml}.
One can observe in Figure \ref{fig:cyclic_mvu_ml} (A) that high probabilities for cycles are beneficial 
for ML and that the discount 
which is most beneficial to ML depends on the probability for the cycle. We have seen in Section \ref{sec:RelMLMVU} 
that the Bellman equation enforces the estimator to use all cycle times from $0$ to ``$\infty$'' and
thus in a sense ``overestimates'' the effect of the cycle. Furthermore,
the probability for the cycle is underestimated by ML, i.e. $\expect[\bar p] < p$  (Section \ref{sec:cyclic_ml}), 
which can be seen as a correction for 
the ``overestimate''. 
The parameter estimate is independent of the true probability and the discount. Therefore,  
a parameter must exist which is most beneficial for ML, i.e. 
ML is biased towards this parameter. The experiment suggests that the most beneficial parameter $p$ is
close to $1$, meaning that ML is biased towards systems with high probabilities for cycles.    

In Figure \ref{fig:cyclic_mvu_ml} (B)  the results of the second experiment are shown. 
In this experiment $\gamma=p=0.9$ and the number of paths is varied. One can observe that
the difference between the ML and the MVU estimator is marginal in comparison to the difference 
to the MC estimator. Furthermore, the bias of ML approaches quickly to $0$ and the MVU and the ML
estimator become even more similar.

\begin{figure}[t]
\begin{center}
\setlength{\epsfxsize}{5.7in}
\centerline{\epsfbox{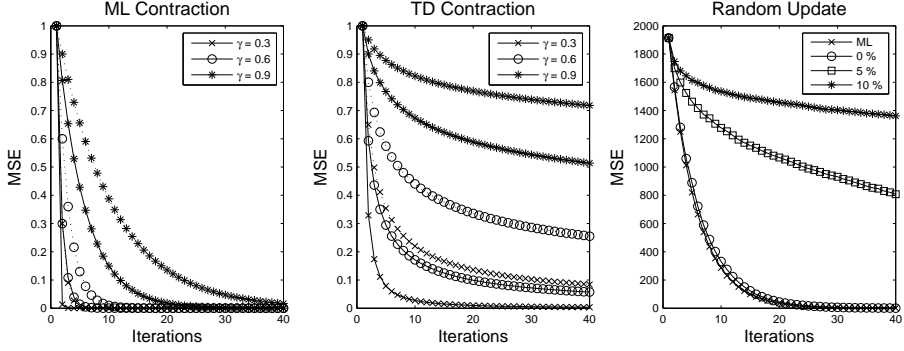}}
\caption{The three plots show the contraction rate of different operators to  the ML solution. The $x$-axis denotes
the number of applications of the operators and the $y$-axis shows the distance to the ML solution. The left 
and the center plot are normed with the initial distance (before the first application). 
\textbf{Left:} The Bellman operator is used. 
The discount $\gamma$ varies from $0.3$ to $0.9$. For each discount value the empirical distance and the
bound (dotted line) is plotted. \textbf{Center:} Same setting as in the left plot but with the ``TD(0)'' operator. 
\textbf{Right:} In this plot $\gamma = 0.9$. The ML curve corresponds again to the Bellman operator. For the 
other three curves only single states are updated with the Bellman operator, whereas the states which are 
updated are chosen randomly.
The percent values denote the deviation from the uniform prior for the states ($0 \%$ means uniform). For the single 
state curves not one update was performed per iteration but $|\mathbb{S}|$ many.
\label{fig:Contraction_Emp}
}
\end{center}
\end{figure}

\subsection{Contraction: ML, iML and TD(0)}
\label{sec:Contraction_Emp}
In a final set of experiments we compared the contraction factor of different operators. 
We generated randomly transition matrices for a state space size of 100 and applied the different operators.
The results are shown in Figure \ref{fig:Contraction_Emp}. The left plot shows the results for the usual 
Bellman operator and the bound for different discount values. In the middle the TD(0) update equation is used and in the
right plot the Bellman operator is applied state wise, whereas the state is chosen randomly from different priors.
The prior probabilities for states $1, \ldots, n:=|\mathbb{S}|$ are given by: 
$p_1 = (1-c) m, p_2 = (1- c + 1/(n-1))m, \ldots, p_n = (1+c) m$, where $m= 1/n$ (mean) and $c$ denotes the deviation
from the uniform prior. If $c=0$ then we have a uniform distribution. If $c=0.1$ then $p_1=0.9 m, 
p_2 = (0.9 + 1/(n-1))m, \ldots, p_n= 1.1 m$.

While we were not able to proof that TD is in general inferior to ML, respectively to iML the plots suggest this
to be the case for typical MRPs. Especially, the contraction of TD (middle plot) to the ML solution is magnitudes slower
than the contraction using the Bellman operator. The state-wise update reduces the contraction speed further. The right
plot shows 
the difference between the fixed point iteration and the state-wise update with the Bellman operator
(corresponding to iML). The contraction factor of the state-wise update depends crucially on
the distribution of visits of the different states. At best (i.e. uniform distribution over the states)
the contraction is about $|\mathbb{S}|$-times slower than the contraction with the Bellman operator applied
to the full state vector.

\section{Summary}
\label{sec:summary-future}
In this work we derived the MVU and compared it to 
different value estimators. In particular, we analyzed the relation between the MVU and the ML estimator. 
It turned out that 
the relation between these estimators is directly linked to the relation between the class of unbiased estimators
 and Bellman estimators. If the ML estimator
 is unbiased then it is equivalent to the MVU and more generally 
the difference between the estimators depends on the bias of ML. This relation is interesting, in particular
as the estimators are based onto two very different algorithms and proving equivalence using combinatorial 
arguments is a challenging task. Furthermore, we demonstrated in this paper that the MC estimator is equivalent to the 
MVU in the undiscounted case if both estimators have the same amount of information. The relation to
TD is harder to characterize. TD is essentially unbiased in the acyclic case and therefore inferior to
the MVU and the ML estimator in this case. In the cyclic case TD is biased and our tools are not applicable.

We want to conclude the section with open problems. Possibly, the most interesting problem is 
the derivation of an efficient MVU algorithm. The combinatorial problems that must be solved appear to be formidable.
Therefore, it is astonishing
that in the undiscounted case the calculation essentially boils down to calculating
the ML estimate. In particular, the exponential runtime of a brute force MVU algorithm which is intractable even 
for simple MRPs decreases in this case to an O$(n^3)$ factor. This efficiency is mainly due to the irrelevance of
the time at which a reward is observed. In the discounted case the time of an observation matters and the 
algorithmical difficulties increase considerably. Instead of the full geometric series of ML with arbitrary long paths
it seems to be needed to make a cutoff at a maximum number of cycles, i.e. 
replacing $(\mathbf{I} - \gamma \mathbf{\bar P})^{-1}$ with something like 
$(\mathbf{I} - \gamma \mathbf{\bar P}^s) (\mathbf{I} - \gamma \mathbf{\bar P})^{-1}$. Yet, Equation \ref{eq:MVU_Simple} 
shows that a weighting factor is associated with each time step and the MVU equation is not that simple.

Another interesting question concerns the bias of the ML estimator. We showed that the normalizations $\{N_i\geq 1\}$ are 
the reason for the bias. Furthermore, if the Full Information Criterion applies then the normalization problem 
is not present and we used a theorem from \citep{SING96} to deduce unbiasedness of ML for this case. Yet, there 
seems to be a deeper reason for the unbiasedness of the ML estimator and the theorem from \citep{SING96} 
appears to be an implication from this and from Corollary \ref{MCIPEqual} (MVU=MC).

\subsection{Discussion}
In the discussion section we address two questions: (1) What is the convergence speed
of the MVU? (2) Which estimator is to be preferred in which setting?
In this section the emphasis is put onto gaining intuition and not on mathematical rigor.

\paragraph{Convergence Speed}
We are interested in the MSE and in the small deviation probability of the MVU.
First, let us state the variance and the Bernstein inequality (e.g. \citep{LUG06}) 
for the first-visit MC estimator with $n$ paths available
for estimation:
\begin{gather*}
\mathbbm{V}[\bar V^{(n)}] = \mse[\bar V^{(n)}] = \frac{1}{n} \mathbbm{V}[R] \\
\end{gather*}
and
\begin{gather*}
\mathbbm{P}(|\bar V^{(n)} - V| \geq \epsilon) \leq 2 \exp\left(- \frac{\epsilon^2 n}{2 \mathbbm{V}[R] + 2 d \epsilon/3}  \right),
\end{gather*}
where
$\mathbbm{V}[R]$ is the variance in the cumulated reward 
(see \citep{SOBEL82} for the variance of a MRP) and $d$ is an upper bound for the cumulated reward of any path,
 i.e. $|\sum R_t - V| < d$. 

How about the MVU? In the undiscounted case the MVU has the same variance 
and small deviation probability if the Full Information Criterion applies. The quality increases 
with further paths into the graph of successor states. Intuitively, the improvement in quality depends on
the "distance'' of the entry point $s'$ in the successor state graph to the state $s$ of which we want 
to estimate the value. A natural distance measure for this setting is the probability to move from state $s$ to $s'$. 
Furthermore, the
improvement will depend on the variation in the cumulative reward of paths starting in $s'$. Paths, that
run through regions in which the reward has high variance will yield a better performance increase than
paths which run through near deterministic regions. The performance will, however, be
lower bounded by the case that all of these $N$ paths start directly in $s$. Therefore, for undiscounted
MRPs the rough lower bound $(1/N) \mathbbm{V}[R]$ will hold:
\begin{gather*}
\frac{1}{N} \mathbbm{V}[R] \leq \mse[\expect[\bar V^{(n)}|\sS]] \leq  \frac{1}{n} \mathbbm{V}[R].
\end{gather*}
If starts in the successor graph are $c$ times more often than starts
in s, i.e. $N=c n$ then 
\begin{gather*}
 \frac{1}{c} \mse[\bar V^{(n)}] \approx   \mse[\expect[\bar V^{(n)}|\sS]]. 
\end{gather*}
Similarly, a ``reasonable'' Bernstein bound of the small deviation probability will lie between
\begin{gather*}
2  \exp\left(- \frac{c \epsilon^2 n}{2 \mathbbm{V}[R]
+ 2 d \epsilon/3} 
\right)
\text{\quad  and \quad}
2 \exp\left(-\frac{\epsilon^2 n}{2 \mathbbm{V}[R]
+ 2 d \epsilon/3}\right).  
\end{gather*}

\paragraph{Choosing an Estimator}
Our study shows that we have essentially a tradeoff between computation time and convergence speed per 
sample. As one would expect, the methods which converge faster have a higher computation time. 
It seems that the fast methods with bad convergence speed are superior if we consider pure computation
time (Experiment 3, Section \ref{exp:comp_time}). However, if there are costs involved for producing 
examples, then the expansive methods become competitive. In a high cost scenario it currently seems best 
to choose the ML/LSTD estimator. The MVU might become an alternative, but an efficient algorithm
is currently missing. Furthermore, the algorithmic problems restricted the numerical comparison to ML and it
is unclear in which setting which estimator is superior.

\appendix
\section{Notation}
\label{sec:notation}
\begin{tabular}{ll}
\hline
$\mu_{ss'}$ & The number of direct transitions from state $s$ to $s'$. \\
\hline
$\pi$ & Path. \\
\hline 
$\pi_i$ & ith state in the path. \\
\hline 
$\Pi_{ss'}$ & Set of all paths from state $s$ to $s'$. \\
\hline
$\mat{\Pi}(\sS)$ & Set of paths that are consistent with $\sS$. \\
\hline
$\prob,\expect,\var$ & Probability measure, expectation and variance. \\
\hline
$H_s$ & Sum of the reward received through transitions from state $s$. \\
\hline
$K_s$ & Number of visits of state $s$. \\
\hline
$\bar p_{ss'}$ & Estimate of the probability for a direct transition from state $s$ to $s'$. \\
\hline
$\bar P_{ss'}$ & Estimate of the transition probability from state $s$ to $s'$. \\
\hline
$\bar R_s$ & Estimator of the reward received through transitions from state $s$. \\
\hline
$\mathbb{S}$ & State space. \\
\hline
$\sS, \sT$ & Sufficient Statistics. \\
\hline
$V_s$ & True value of state $s$. \\
\hline
$\bar V_s$ & Estimated value of state $s$. Concrete estimator is section dependent. \\
\hline
$\bar V_s^{(i)}, \bar P_{ss'}^{(i)}, \ldots$ & Superscripts denote values after the ith run. \\
\hline
$\mat{V}, \mat{P}, \ldots$ & Vectors and Matrices. \\
\hline
\end{tabular}

\section{Unbiased TD($\lambda$)}
\label{sec:unbiased-td0-version}
In this section we introduce a (minorly) modified TD($\lambda$) estimator. The estimates 
are, in contrast to the standard TD($\lambda$) estimator, independent of the initialization.
In the acyclic case this is already enough to guarantee unbiasedness of TD($\lambda$).
We first discuss the TD(0) case. This case contains the major arguments in an 
accessible form.

\subsection{TD(0)}
We first restate the TD(0) equation through unfolding the recursive definition                                
(eq. \ref{TD0Def}, p. \pageref{TD0Def}).
\begin{lemma}
\label{SumFormTD}
If the TD(0) estimator is initialized with 0
then for an acyclic MRP it equals 
$$ \bar V^{(n)}_s = \sum_{i=1}^n \beta_i R^{(i)} +
\sum_{s' \in \mathbb{S}} \Bigl( \sum_{i=1}^n T_{i,s'} \beta_i \gamma \bar V^{(i-1)}_{s'} \Bigr), $$
where $\beta_i:= \Bigl(\alpha_i \prod_{j=i+1}^{n} (1- \alpha_j)  \Bigr) $, 
$R^{(i)}$ is the received reward in path $i$ and $T_{i,s'}$ a random variable which is one if in run $i$ the state $s'$
followed upon state $s$ and is zero otherwise. 
\end{lemma}
\begin{proof}
The recursive TD(0) definition 
(eq. \ref{TD0Def}) can be written as:
$
\bar V_s^{(n)} = \bar V^{(n-1)}_{s} 
(1- \alpha_n) +  \alpha_n ( R^{(n)} +  \gamma \bar V^{(n-1)}_{s'}). 
$
Substituting $\bar V^{(n-1)}_{s}$:
\begin{align*}
&\bar V_s^{(n)} = 
\bigl(\gamma \bar V^{(n-2)}_{s} (1- \alpha_{n-1}) +  \alpha_{n-1} ( R^{(n-1)} +  \gamma \bar V^{(n-2)}_{s''})\bigr) 
(1- \alpha_n) +  \alpha_n ( R^{(n)} +  \gamma \bar V^{(n-1)}_{s'}) \\
&= \gamma \bar V^{(n-2)}_{s} (1- \alpha_{n-1})(1-\alpha_n) +  \alpha_{n-1}(1-\alpha_n) ( R^{(n-1)} +  \gamma \bar V^{(n-2)}_{s''})
+ \alpha_n ( R^{(n)} +  \gamma \bar V^{(n-1)}_{s'}) \\
&= \ldots = \sum_{i=1}^n  
\Bigl(\alpha_i \cdot \prod_{j=i+1}^{n} (1- \alpha_j)  \Bigr)  
\left(R^{(i)} + \gamma \bar V^{(i-1)}_{s^{(i-1)}}\right) 
=:  \sum_{i=1}^n \beta_i \left(R^{(i)} + \gamma \bar V^{(i-1)}_{s^{(i-1)}}\right). 
\end{align*} 
\end{proof}
The estimate contains the values $\bar V_{s'}^{(0)}$ which bias the estimator towards the initialization.
The estimator can be made unbiased for acyclic MRPs 
by excluding these values and by guarantying that the $\beta_i$ sum to one.
Modification \ref{TD0Mod} does exactly this\footnote{The modification is easy to implement with a 
$TD(\lambda)$ algorithm. $\lambda$ is  
set to 0 if the successor state is initialized and it is set to 1 if the successor state is not. 
Further, the initial learning rate must be 1.}. 

\begin{algorithm}[h!]
\floatname{algorithm}{Modification}
\caption{Modified TD(0)}
\label{TD0Mod}
\begin{algorithmic}
\IF {$\bar V^{(i)}_s$ has seen no example} 
\STATE set the learning rate for this step to 1.
\ENDIF
\IF {$\bar V^{(i)}_{s'}$ has seen no example} 
\STATE first update the estimate $\bar V^{(i)}_{s'}$
\ENDIF
\STATE Use the TD(0) update rule (\ref{TD0Def}). 
\end{algorithmic}
\end{algorithm}

Setting the learning rate for the first example to 1 eliminates the initialization of $\bar V_s$. The second
rule assures that the initialization of the estimators of the successor states is eliminated. 
Setting the learning rate $\alpha_1$ to 1 has also the effect that the weighting factors $\beta_i$ 
sum to one, independent of the learning rate. For example for n=3, we have
$ \sum_{i=1}^3 \beta_i =  1 ( 1- \alpha_2) (1-\alpha_3) + \alpha_2 (1 - \alpha_3) + \alpha_3 = 1. $

\begin{theorem}
The modified TD(0) estimator 
is unbiased if the MRP is acyclic. 
\end{theorem}
\begin{proof}
We prove this by induction. 
We start with the terminal states for which $\bar V_s = 0  = V_s$ holds. The induction step considers now the states 
which have only successors that have already been handled. This way the complete state space will be addressed. 
The expectation has the form (Lemma \ref{SumFormTD}):
\begin{align*}
&\mathbbm{E}\Bigl[
\sum_{i=1}^n \beta_i R^{(i)} +
\sum_{s' \in \mathbb{S}} 
\Bigl( \sum_{i=1}^n T_{i,s'} \beta_i \gamma \bar V_{s'}^{(i-1)} \Bigr) 
\Bigl|K_s=n\Bigr] = \\
& \mathbbm{E}\Bigl[ \sum_{i=1}^n \beta_i R^{(i)} \Bigl|K_s=n\Bigr]  + 
\sum_{s' \in \mathbb{S}} \sum_{i=1}^n \beta_i \gamma  \mathbbm{E}\Bigl[T_{i,s'}  \bar V_{s'}^{(i-1)}  \Bigl|K_s=n\Bigr]  = \\
& \mathbbm{E}\Bigl[ \sum_{i=1}^n \beta_i R^{(i)} \Bigl|K_s=n\Bigr]  + 
\sum_{s' \in \mathbb{S}} p_{ss'} \sum_{i=1}^n \beta_i \gamma  \mathbbm{E}\Bigl[  \bar V_{s'}^{(i-1)}  \Bigl|K_s=n\Bigr].  
\end{align*}
It remains to show that  $\mathbbm{E}\Bigl[ \bar V_{s'}^{(i-1)}\Bigl|K_s=n\Bigr]$ is unbiased. For
$i\geq 2$ this follows from the induction hypothesis. For the case $i=1$ Modification \ref{TD0Mod} guarantees that the
estimator has at least one example for estimation and is  unbiased due to the induction hypothesis.
Furthermore, the $\beta_i$'s sum to one due to the modification and
$ \sum_{s' \in \mathbb{S}} p_{ss'} \gamma    V_{s'} \sum_{i=1}^n \beta_i  = \sum_{s' \in \mathbb{S}} p_{ss'} \gamma V_{s'}$.
\end{proof}

\subsection{TD($\lambda$)}
The TD($\lambda$) case is essentially the same. The main difference is that the estimates
of all states of a path are used. Therefore, it is not enough that the estimators of the 
direct successor states are set to ``reasonable'' values, but all states of the path must be:
\begin{algorithm}[h!]
\floatname{algorithm}{Modification}
\caption{Modified TD($\lambda$)}
\label{TDLMod}
\begin{algorithmic}
\IF {$\bar V^{(i)}_s$ has seen no example} 
\STATE set the learning rate for this step to 1.
\ENDIF
\IF {for a successor $s'$ in the path $\bar V^{(i)}_{s'}$ has seen no example} 
\STATE first update the estimate $\bar V^{(i)}_{s'}$
\ENDIF
\STATE Use the TD($\lambda$) update rule. 
\end{algorithmic}
\end{algorithm}

\begin{theorem}
\label{th:TDlam_unbiased}
The modified TD($\lambda$) estimator is unbiased if the MRP is acyclic.
\end{theorem}
\begin{proof}
Proof by induction. 
\textbf{Induction Hypothesis:} $\mathbbm{E}[\bar V_s|K_s =n] = V_s.$
\\
\textbf{Induction Basis:} For terminal states the Hypothesis trivialy holds.
\\
\textbf{Induction Step:} Let  $\pi(i,j)$ be state $i$ 
in path $j$ and let $R_{\pi(i,n)}$ be the reward received at state $j$ in run $i$.
In the acyclic case TD($\lambda$) can be written as

\begin{align*}
\bar V_s^{(n)} &=  (1-\alpha_{n}) \bar V_s^{(n-1)}   + \alpha_{n} \left( 
\sum_{i=1} (\gamma \lambda)^{i-1} R_{\pi(i,n)} + \gamma (1-\lambda) \sum_{i=2} (\gamma \lambda)^{i-2} 
\bar V_{\pi(i,n)} \right) \\
&= (1- \alpha_{1}) \bar V_s^{(0)} +  \sum_{j=1}^n
\beta_{j} \left( 
\sum_{i=1} (\gamma \lambda)^{i-1} R_{\pi(i,j)} + \gamma (1-\lambda) \sum_{i=2} (\gamma \lambda)^{i-2} 
\bar V_{\pi(i,j)} \right). 
\end{align*}
We suppressed the ``iteration'' index of $\bar V_{\pi(i,j)}$ for readability. Like in the TD(0) case
$\beta_j:=  \Bigl(\alpha_j  \prod_{k=j+1}^{n} (1- \alpha_k)  \Bigr)$.
Applying the expectation operator and using $\alpha_{1} = 1$, we get
\begin{align}
&\expect[\bar V_s^{(n)}|K_s = n]
= \expect\left[\sum_{j=1}^n \left.
\beta_{j} \left( 
\sum_{i=1} (\gamma \lambda)^{i-1} R_{\pi(i,j)} + \gamma (1-\lambda) \sum_{i=2} (\gamma \lambda)^{i-2} 
\bar V_{\pi(i,n)} \right) \right| K_s =n \right]  \notag \\
&=  \sum_{j=1}^n \beta_{j}  \left( 
 \sum_{i=1} (\gamma \lambda)^{i-1} \expect[R_{\pi(i,j)} | K_s =n ] 
+ \gamma (1-\lambda)   \sum_{i=2} (\gamma \lambda)^{i-2} 
\expect[\bar V_{\pi(i,j)}  | K_s =n ] \right).
\label{eq:td_lambda_helper1}
\end{align}
Instead of $\expect[R_{\pi(i,j)}]$ and $\expect[\bar V_{\pi(i,j)}]$ we 
use $\expect[R_i]$ and $\expect[\bar V_i]$ in the following 
to denote the expected reward in 
step $i$, respectively the expected value estimate in step $i$ (expected state times expected value estimate
for that state). Due to the induction hypothesis 
$$\expect[\bar V_i  | K_s =n ]  = V_i = \sum_{j=i} \gamma^{j-i} \expect[R_j].$$
Substituting this term into equation \ref{eq:td_lambda_helper1}:
\begin{align*}
& \sum_{j=1}^n \beta_{j}  \left( 
 \sum_{i=1} (\gamma \lambda)^{i-1} \expect[R_i] 
+ \gamma (1-\lambda)   \sum_{i=2} (\gamma \lambda)^{i-2} 
\sum_{j=i} \gamma^{j-i} \expect[R_j] \right).
\end{align*}
Taking a specific $\expect[R_i]$, we see that for the coefficient 
\begin{align*}
&(\gamma \lambda)^{i-1} + \gamma(1-\lambda)(\gamma^{i-2} \lambda^{i-2} + \ldots + \gamma^{i-2} \lambda + 1) \\
&= \gamma^{i-1} \left( \lambda^{i-1} + (1-\lambda) \sum_{j=0}^{i-2} \lambda^j\right) = 
 \gamma^{i-1} \left( \lambda^{i-1} + (1-\lambda) \frac{1 - \lambda^{i-1}}{1-\lambda}\right)  = \gamma^{i-1}
\end{align*}
holds. We know already that the $\beta_j$ sum to one. Hence, the modified TD($\lambda$) is unbiased.
\end{proof}

\section{Proofs}
\label{sec:proofs}

\subsection{Unbiased Estimators - Bellman Equation}
\textbf{Lemma \ref{lem:NormalizationBiased}}
\textit{
For the MRP from Figure \ref{fig:cyclic_example} (B) there exists no parameter estimator $\bar p$
such that $V_i(\bar p)$ is unbiased for all states $i$.}
\label{proof:NormalizationBiased}
\begin{proof}
Assume that $\bar V_1,\bar V_2$ are unbiased, i.e. $\expect[\bar V_1|\{N_1\geq 1\}] = \expect[\bar V_2] = V_1 = V_2$ and
the estimator fulfills the Bellman equation, i.e.  $\bar V_1 = \bar V_2$ on $N_1 : =\{N_1 \geq 1\}$. Then
\begin{gather*}
  \expect[\bar V_2|N_1]    \overset{\text{Bellm.}}{=} \expect[\bar V_1|N_1]  \overset{\text{unb.}}{=} V_1 
  \overset{\text{Bellm.}}{=} V_2  \overset{\text{unb.}}{=} \expect[\bar V_2]. 
\end{gather*}
We used for the first equality that $\bar p_{12}$ must equal 1 as there is only one connection leading away from
state 1. The derived equality shows that the average value estimate  $\bar V_2$ must be the same 
as the average estimate for the case that the connection $2\rightarrow 1$ has been taken at least once.
This implies that the value estimate for the case that only the connection $2 \rightarrow 3$ has been taken
must be the same as the average value estimate:
\begin{gather*}
\expect[\bar V_2]   =  \expect[\bar V_2|N_1] \prob[N_1]  +   \expect[\bar V_2|N_1^c] \prob[N_1^c] 
\overset{\text{unb.}}{=} \expect[\bar V_2] \prob[N_1]  +   \expect[\bar V_2|N_1^c] \prob[N_1^c], 
\end{gather*}
where $N_1^c$ denotes the event $N_1 = 0$. This implies that 
$\expect[\bar V_2|N_1]  = \expect[\bar V_2]  =\expect[\bar V_2|N_1^c]$.

There are two possibilities to achieve this equality: (1) All three terms are $0$. In particular, 
$\expect[\bar V_2] = 0 \not = V_2$. This contradicts unbiasedness. (2)$\expect[\bar V_2|N_1^c] \not = 0$:
As $R_{23} = 0$ that means that $\bar p_{21} \not = 0$, despite the fact that this transition has not been observed.
Furthermore, this implies that $\bar p_{12} = 1$ as otherwise no valid MRP is defined. As a consequence 
$\bar V_1 = \bar V_2$ on both the events $N_1$ and $N_1^c$, in particular 
$\expect[\bar V_1|N_1^c] = \expect[\bar V_2|N_1^c] \not = 0$. Now, we get a contradiction with the following argument:
$$ V_1 + \expect[\bar V_1|N_1^c] \overset{\text{unb.}}{=} \expect[V_1|N_1] + \expect[\bar V_1|N_1^c] = 
\expect[V_2|N_1] + \expect[\bar V_2|N_1^c] = \expect[\bar V_2] \overset{\text{unb.}}{=} V_2 
\overset{\text{Bellm.}}{=} V_1,
$$
as the equality can only hold if $\expect[\bar V_1|N_1^c] = 0$.
\end{proof}

\textbf{Lemma \ref{lem:ValueFuncBias}}
\textit{
For the MRP from Figure \ref{fig:cyclic_example} (A) and for $n=1$ there exists no parameter estimator $\bar p$
that is independent of $\gamma$ such that $V(\bar p)$ is unbiased for all parameters $p$ and all
discounts $\gamma$.}
\label{proof:ValueFuncBias}
\begin{proof}
For $V(\bar p)$ to be  unbiased, it must hold that 
\begin{align*}
\expect\left[(1-\bar p)\sum_{i=0}^\infty \gamma^i \bar p^i \right] =
(1-p) \sum_{i=0}^\infty  \gamma^i p^i   \Rightarrow
\sum_{i=0}^\infty \gamma^i  (\expect[(1-\bar p )\bar p^i] - (1-p)p^i)   = 0.
\end{align*}
If the equality holds for all $\gamma\in (0,1)$, then $\expect[(1-\bar p)\bar p^i] = (1-p)p^i$ for $i\geq 0$. 
Otherwise, 
with $x_i := \expect[(1-\bar p )\bar p^i] - (1-p)p^i$ and $x_n$ being the first term different from $0$ ($|x_n|> 0$): $|\gamma^n x_n| = |\sum_{i=n+1}^\infty \gamma^i x_i|$ and  therefore
$|x_n| = |\gamma \sum_{i=n+1}^\infty \gamma^{i-(n+1)} x_i|$. We can now adjust the discount 
to downscale the right hand side arbitrary low, while the left side stays unaffected.
The sequence $|x_i|$ is bounded, i.e. $|x_i| \leq \max\{\expect[(1-\bar p)\bar p^i], (1-p)p^i \}$ 
for all $i$. Futhermore, both terms are bounded by $\max_{a\in[0,1]} (1-a)a^i$. The maximum is 
reached for $a=i/(i+1)$ and the maximal value over all $i$ is reached for $i=0$. The value for $i=0$ is
$1$. Therefore,
\begin{gather*}
|\gamma \sum_{i=n+1}^\infty \gamma^{i-(n+1)} x_i| < \gamma \sum_{i=n+1}^\infty \gamma^{i-(n+1)} 1=
 \frac{\gamma}{(1-\gamma)}.
\end{gather*}
For $\gamma < |x_n|/(1-|x_n|)$ the term and the remaining part of the sum becomes smaller than $|x_n|$.
$|x_n|/(1/4-|x_n|)$ is always larger than $0$ as $|x_n| > 0$ and for $|x_n| \rightarrow 1$ the discount $\gamma$ can be
chosen arbitrary large. 
In summary this contradicts the assumption and $\expect[(1-\bar p)\bar p^i]$  must equal $(1-p)p^i$ for all $i \geq 0$.

Therefore, $\expect[1-\bar p] = 1 - p \Rightarrow \expect[\bar p ] =  p$ and 
$\expect[(1-\bar p) \bar p ] = (1 - p) p \Rightarrow \expect[\bar p^2 ] =  p^2$. 
Consequently, $\bar p$ must be a constant. Otherwise, we get a contradiction with the following argument:
The possible values of $\bar p$ are countable (countable many outcomes). We denote the values with $a_i$ and
with $q_i$ the probabilities for the values $a_i$. From $\expect[\bar p^2] = \expect[\bar p]^2$ it follows that  
$\sum_{i=0}^\infty q_i a_i^2  = \sum_{i=0}^\infty \sum_{j=0}^\infty  q_i q_j a_i a_j  \Rightarrow 
\sum_{i=0}^\infty \sum_{j\not=i}  q_i q_j a_i a_j   = 0 $. Furthermore, $q_i, a_i \geq 0$ for all $i$ and therefore 
$q_i q_j a_i a_j   = 0 $ for all $i\not=j$. As a consequence, there can be only one $a_i> 0$ with $q_i > 0$. 
Because $\expect[\bar p] = p$ it holds that $a_i = p/q_i$ and because $\expect[\bar p^2] = p^2$ it holds that 
$a_i = p/\sqrt{q_i}$. Hence, $q_i= 1$ and the parameter estimate is almost surely a constant $p$. Hence, for a MRP 
with $\tilde p = p/2$ the estimator will not be unbiased.
\end{proof}

\subsection{Markov Reward Process}
\begin{lemma} 
\label{lem:complete}
A MRP with finite state space and iid sequences forms an s-dimensional exponential family, where s is the number of free 
MRP parameters. 
\end{lemma}
\begin{proof}
Firstly, we demonstrate
 that the transition distribution forms an exponential family. The density can be written as
\begin{align*}
\prob(X_1 = \pi^{(1)}, \ldots, X_l = \pi^{(l)}) = \prod_{i=1}^\infty P_{\pi_i}^{c_i}  
= \exp\Bigl( \sum_{i=1}^\infty c_i \log P_{\pi_i}  \Bigr), 
\end{align*}
with $\pi^{(i)}$ being the observed paths, $(\pi_i)_{i \in \mathbbm{N}}$ the set of paths,
 $c_i$ the number of times path $i$ has occurred and $P_\pi$ the probability of
 path $\pi$. The parameters $P_\pi$ are redundant.
We explore now the MRP structure to find natural parameters that are not functionally dependent. 
The size of this set of parameters is the number of necessary MRP parameters, that is 
$$\sharp\text{Starting States} -1 + \sum_{i \in \statespace} \bigl(\sharp \text{Direct Successors of } i -1\bigr).$$
We reformulate the exponential expression to reduce the number of parameters. First, one can observe that
$\prod_{i=1}^\infty P_{\pi_i}^{c_i}$ is equivalent to $\prod_{i \in \mathbb{S}} \left( p_i^{n_i} \prod_{j \in \mathbb{S}} 
p_{ij}^{\mu_{ij}} \right)$,
where $n_i$ is the number of starts in state $i$. The parameters are still redundant: Let state $1$ 
be a starting state and $S$ the remaining set of starting states, then 
$p_1 = 1- \sum_{j \in S}  p_j$.
Furthermore, we have one redundant parameter $p_{ij}$ for every state $i$. The first problem 
can be overcome by using $A(\theta)$ in the following way:
$n \log \left( 1- \sum_{i\in S} p_i \right) + \sum_{i \in S} n_i \log \frac{p_i}{\left(1 - \sum_{j\in S} p_j \right)}$.
Here, $A(\theta)$ equals the $n$ term and $n_i$ is the number of starts in state $i$. Using the same approach
for the transition parameters results in 
$K_i \log \left( 1- \sum_{j\in S(i)} p_{ij} \right) 
+ \sum_{j \in S} \mu_{ij} \log \frac{p_{ij}}{\left(1 - \sum_{u\in S(i)} p_{iu} \right)}$, with
$S(i)$ being the set of successor states of  $i$ without the first successor.
This time the $K_i$ term cannot be moved into $A(\theta)$, as $K_i$ is data dependent. 
This problem can be overcome by observing that $K_i = n_i + \sum_{j\in \mathbb{S}} \mu_{ji}$ and by 
splitting the $K_i$ terms. As a result we get 
\begin{align*}
&\exp\Biggl(n \log\left(1- \sum_{i \in S} p_i\right) \left(1-\sum_{u \in S(1)} p_{1u} \right) + 
\sum_{i \in S} n_i \log \frac{p_i \left(1-\sum_{u \in S(i)} p_{iu} \right)}{\left(1-\sum_{j\in S} p_j \right)}  \\
&+ \sum_{i} \sum_{j\in S(i)} \mu_{ij} \log \frac{p_{ij} \left(1-\sum_{u \in S(j)} 
p_{ju}\right) }{\left(1-\sum_{u \in S(i)} p_{iu} \right)} \Biggr).
\end{align*}
If the reward is deterministic and the examples consist of state sequences, then the MRP forms an exponential family. 
If the reward is a random variable then it depends on the distribution of this random variable. 
In many cases, like for the binomial or multinomial distribution, the resulting MRP still forms an exponential family.
\end{proof}

\subsection{MVU}
\label{app:MVU}
\textbf{Theorem \ref{th:MVU_CONV}}
\textit{
$\expect[\bar V|\sS]$ converges on average to the true value. Furthermore, it converges almost surely 
if the MC value estimate is upper bounded by a random variable $Y \in L^1$.
}
\begin{proof}
The estimator converges on average, because $\expect[ |\expect[\bar V|\sS] - V|] \leq 
\expect[ |\bar V - V|] \overset{n \rightarrow \infty}{\longrightarrow} 0$, where $n$ denotes the
 number of observed paths and convergence follows from the MC convergence. 
The inequality follows from the Rao-Blackwell Theorem, respectively
from the Jensen inequality because $|\cdot|$ is convex.

We need to show that 
$$\lim_{n\rightarrow \infty} \expect[\bar V|\sS] = V \text{ \quad a.s.},$$ 
where $n$ denotes again the number of observed paths for almost sure convergence.

We use a statement from \citep{BAUER95}[§15 Conditional Expectation, (15.14)], which 
says that  $\lim_{n\rightarrow \infty} \expect[\bar V|\sS] = V \text{ \quad a.s.}$ if
$\bar V$ converges almost surely and if it is upper bounded by a random variable $Y \in L^1$.
The upper bound comes from the Lebesgue convergence Theorem and the statement uses 
that for the conditional expectation if holds that
 $ \lim \bar V = V \text{ a.s. }$ implies  $\expect[\lim \bar V|\sS] = V$ a.s..
\end{proof}

\subsection{ML Estimator}
\label{sec:proof_ml}
\noindent
\textbf{Theorem \ref{theo:LSTD_unbiased}} \textit{
The ML estimator is unbiased if the MRP is acyclic. 
}
\begin{proof}
The value function can be written as
$$  V_s  = \mathbbm{E}[R_{s}] + \sum_{s'\in \mathbb{S}} \sum_{\pi \in \Pi_{ss'}} P_{\pi} \gamma^{|\pi|} 
\mathbbm{E}[R_{s'}], $$
where  $\Pi_{ss'}$ is the set of paths from $s$ to $s'$, $P_{\pi}$ the probability of path $\pi$, $|\pi|$ the 
length of the path and $\expect[R_s] = \sum_{s' \in \mathbb{S}} p_{ss'} \expect[R_{ss'}]$.
The ML estimator can be written in the same form, whereas $P_\pi$ is replaced with $\bar P_\pi
:= \prod_i \bar p_{\pi_{i}\pi_{i+1}}$ and the expected reward with the reward estimator.
The sample mean estimator $\bar p$ is unbiased and 
the reward estimator is unbiased because of our initial assumption.
The main problem
is to show that $\bar P_{\pi}$ is unbiased, i.e. that
\begin{align*}
&
\mathbbm{E}\Bigl[\prod_{i}  \bar p_{\pi_{i}\pi_{i+1}}\Bigr|K_s=n\Bigr] 
\overset{?}{=} \prod_{i}  p_{\pi_{i}\pi_{i+1}}.
\end{align*}
The last of these  estimators (denote it with $p_{\hat s \hat{ \hat s}}$) is conditionally independent of the 
others given the number of visits of state $\hat s$ ($K_{\hat s}$). This is also the main point where acyclicity is needed.
Using this together with the law of total probability and the fact that $\bar p$ is unbiased, leads to the 
following statement (with $L$  being the length of the path $\pi$):
\begin{align*}
&\mathbbm{E}\Bigl[\prod_{i=1}^{L-1}  \bar p_{\pi_{i}\pi_{i+1}}\Bigr|K_s=n\Bigr]  = 
\sum_{l=1}^n \mathbbm{E}\Bigl[\prod_{i=1}^{L-1}  
\bar p_{\pi_{i} \pi_{i+1}}\Bigr|K_s=n, K_{\hat s}=l\Bigr] \mathbbm{P}[K_{\hat s}=l|K_s=n]  \notag \\
&\overset{ind}{=} 
\sum_{l=1}^n \mathbbm{E}\Bigl[\prod_{i=1}^{L-2}  \bar p_{\pi_{i} \pi_{i+1}}\Bigr|K_s=n, K_{\hat s}=l\Bigr] 
p_{\hat s \hat{ \hat s}}
\mathbbm{P}[K_{\hat s}=l|K_s=n]  \notag \\
&=p_{\hat s \hat{ \hat s}} \sum_{l=1}^n \mathbbm{E}\Bigl[\prod_{i=1}^{L-2}  
\bar p_{\pi_{i} \pi_{i+1}}\Bigr|K_s=n, K_{\hat s}=l\Bigr] 
\mathbbm{P}[K_{\hat s}=l|K_s=n] =  
p_{\hat s \hat{ \hat s}} \mathbbm{E}\Bigl[\prod_{i=1}^{L-2}  \bar p_{\pi_{i} \pi_{i+1}}\Bigr|K_s=n\Bigr]. 
\label{eq:unbiased}
\end{align*}
We used that for $l=0$ the last estimator $\bar p$ in the product is zero. The 
procedure has to be repeated for every $\bar p$. As a result 
the expectation of this estimator is equal to the path probability. One can handle the reward estimator with the same 
procedure. In summary we find that the value estimator is unbiased.
\end{proof}

\section{Counterexamples}
\subsection{MC - TD}
\label{Counter:MC_TD}
\begin{figure}[h]
\begin{center}
\setlength{\epsfxsize}{5.in}
\centerline{\hspace{1cm} \epsfbox{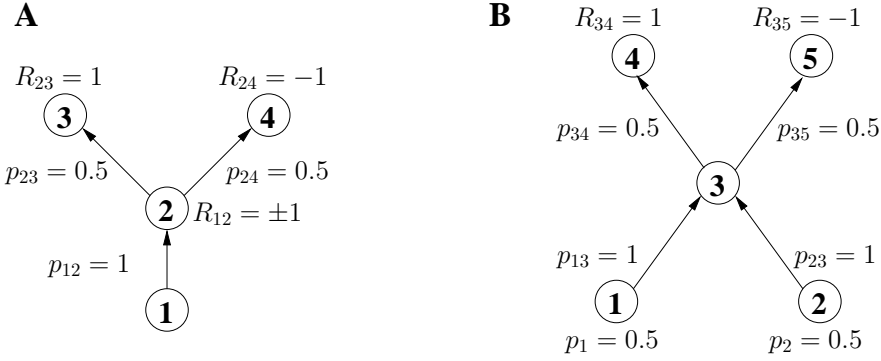}}
\caption{\textbf{A}: A MRP for which TD is inferior to MC. 
The transition from state 1 to state 2 is followed by a reward $R_{12}=+1$ and $R_{12}=-1$ with probability $p=0.5$ each.
\textbf{B}: A MRP for which MC is inferior to TD. 
No reward is received for transitions $1 \rightarrow  2$ and $1\rightarrow 3$.
$p_1$ and $p_2$ are the probabilities to start in state 1 and 2. }
\label{CounterEx}
\end{center}
\end{figure} 

We present two examples in this section. In the first example MC has a lower MSE than TD(0) and is at least as good as 
TD($\lambda$) for every $\lambda$. In the second example TD(0) is superior to MC.

\subsubsection{MC Superior to TD}
\label{CounterTD}
Figure \ref{CounterEx} (A) shows an example for which the MC estimator is superior. 
We assume that the learning rate $\alpha_i$ of TD(0) is between 0 and 1, 
that the learning rate in the first step is 1 ($\alpha_1=1$) and that the estimator is initialized 
to 0 (we use this assumption for readability, it is also possible to use the unbiased TD(0) estimator 
(Modification \ref{TD0Mod})). 
Let state 1 be the starting state, $n$ be the number of observed paths 
and let $\gamma=1$ for simplicity.

The MC estimator for state 2 is $1/n \sum_{i=1}^n Y_i$, where $Y_i = R_{23}$ or $Y_i = R_{24}$ are the rewards
received after a transition from state 2 to state 3 or 4.
For state 1 we obtain 
$1/n \sum_{i=1}^n (Y_i + R_{12}^{(i)})$, where the $Y_i$ are the same as before and $R_{12}^{(i)}$ is 
the received reward after a transition 
to state 2. The MC estimator is a weighted average of the examples and it is the optimal unbiased linear 
estimator (eq. \ref{eq:lin_est}) as $\alpha_i=1/n$ for all $i$. 

We now analyze the TD(0) estimator. Consider two different sequences $\alpha_i$ and $\tilde \alpha_i$, $i=1,\ldots,n$, of 
learning rates for the TD(0) estimators $\bar V_1$ and $\bar V_2$.
The TD(0) estimator $\bar V_2$ can be written as (Lemma \ref{SumFormTD}, Appendix \ref{sec:unbiased-td0-version}) 
\begin{align*} 
&\bar V^{(n)}_2 =  \sum_{i=1}^{n}   \Bigl(\tilde \alpha_i  \prod_{j=i+1}^{n} (1- \tilde \alpha_j)  \Bigr) Y_i =: 
\sum_{i=1}^n \tilde \beta_i Y_i. 
\end{align*}
The estimator is unbiased and has minimal variance if and only if $\tilde \beta_i = 1/n$. This 
can be enforced by choosing $\tilde \alpha_i=1/i$.
For state 1 we obtain
\begin{align}
\bar V^{(n)}_1 
&=\sum_{i=1}^n \beta_i (\bar V^{(i-1)}_2 + R_{12}^{(i)}) = 
\sum_{i=1}^n \beta_i 
\Bigl(\sum_{j=1}^{i-1} \tilde \beta_j Y_j + R_{12}^{(i)}\Bigr) \\
&= \Bigl(\sum_{i=1}^n \beta_i R_{12}^{(i)}\Bigr) +  \Bigl(\sum_{i=1}^{n-1} \bigl(\tilde \beta_i \sum_{j=i+1}^n \beta_j \bigr) Y_i\Bigr)
=: \Bigl(\sum_{i=1}^n \beta_i R_{12}^{(i)}\Bigr) +  \Bigl(\sum_{i=1}^{n-1} \gamma_i Y_i\Bigr), \notag
\end{align}
where $\beta_i = \alpha_i \prod_{j=i+1}^n (1-\alpha_j)$. 
Using the Bienaymé equality (e.g. \citep{BAUER95}) the variance of the estimator takes the following form
\begin{equation*}
 \mathbbm{V}(\bar V^{(n)}_1) \overset{ind}{=}  
\mathbbm{V}(\sum_{i=1}^n \beta_i R_{12}^{(i)}) +  \mathbbm{V}(\sum_{i=1}^{n-1} \gamma_i Y_{i}) 
\overset{iid}{=}  \mathbbm{V}(R_{12}^{(1)})  \sum_{i=1}^n \beta_i^2 
+  \mathbbm{V}(Y_{1}) \sum_{i=1}^{n-1} \gamma_i^2,  
\end{equation*}
where ``ind''  abbreviates  ``independence''.
$Y_1$ and  $R_{12}^{(1)}$ have the same variance. 
With $\gamma_n=0$
$$  \mathbbm{V}(\bar V^{(n)}_1) =   \mathbbm{V}(Y_{1}) (\sum_{i=1}^n \beta_i^2  + \sum_{i=1}^{n} \gamma_i^2). $$  
Because $0\leq \beta_i, \gamma_i \leq 1$ and $\sum_{i=1}^n \beta_i = \sum_{i=1}^n \gamma_i =1$ 
(see Appendix \ref{sec:unbiased-td0-version}) 
this term would be minimal if and only if $\beta_i = \gamma_i = 1/n$.
From $\beta_i = \tilde \beta_i = 1/n$, however, it follows that $\gamma_i = 1/n \sum_{i+1}^{n-1} 1/n =
(n-i-2)/n^2 \not = 1/n$. Hence optimality cannot be achieved.
Since both MC and TD are unbiased, we obtain $\mse[MC] < \mse[TD]$.

This example demonstrates a major weakness of TD, namely that it is impossible for TD to weight the 
observed paths equally, even for simple MRPs. Furthermore, MC is for this example the optimal unbiased
value estimator and TD($\lambda$) is unbiased. The optimality of MC is a direct implication of 
Corollary \ref{cor:MC_Opt} from Section \ref{MCRelSec}. Therefore $\mse[MC] \leq \mse[TD(\lambda)]$ for each $\lambda$.

\subsubsection{TD Superior to MC}
\label{CounterMC}
Figure \ref{CounterEx} (B) shows an example where TD(0) is superior.
Let the number of observed paths be $n=2$ and $\gamma =1$. The value of all states is zero. 
TD(0) and MC are unbiased for this example. The variance of the MC estimator 
for states 1,2 and 3 is therefore given by
\begin{align*}
\expect[\bar V_3^2] =& \mathbbm{P}[R^{(1)} = 1, R^{(2)} = 1]  \cdot 1^2 +
\mathbbm{P}[R^{(1)} = -1, R^{(2)} = -1]  \cdot (-1)^2 \\
&+ \mathbbm{P}[R^{(1)} = 1, R^{(2)} = -1]  \cdot 0 + \mathbbm{P}[R^{(1)} = -1, R^{(2)} = 1]  \cdot 0 
\\
=&(1/4) 1^2 + (1/4) (-1)^2  + (2/4) \cdot 0  = 1/2, \\
\expect[\bar V_1^2] =& \expect[\bar V_2^2] = (1/4) 1/2 + (1/2) 1  + 0 =  5/8,
\end{align*}
where $R^{(i)}$ denotes the received reward in run $i$. The first term in the second line results 
from starting two times in state 1 or 2 and the second term 
in the second line from a single start in state 1.
Setting the learning rate $\alpha_i$ to $\alpha_i=\frac{1}{i}$ for TD, the
estimator for state 3 is equivalent to the corresponding MC estimator and therefore the variance is 1/2.  
In the first run the standard TD(0) update rule uses 
the initialization value of state 3 to calculate 
the estimate in state 1 or 2. This is advantageous and results in a variance of
1/2. Without exploiting this advantage the
variance is 17/32. This is still lower than the variance of the MC estimator. Since
both estimators are unbiased we obtain $\mse[TD(0)]<\mse[MC]$.

\subsection{MVU/MC - ML}
\label{sec:counter_mvu_ml}
We show by means of counterexamples that neither the MVU is superior to the ML estimator nor 
is the ML estimator superior to the MVU or to the MC estimator. 
We use again the MRP from Figure \ref{fig:cyclic_example} (A) on page  \pageref{fig:cyclic_example}  
with $n=1$. 
As we showed before, the value for state $1$ is $(1-p)/(1-\gamma p)$ and the ML estimate
is $(1-\bar p)/(1-\gamma \bar p)$, where $\bar p = i/(i+1)$  and $i$ denotes the number of times
the cyclic connection has been taken. The MC estimate and therefore the MVU estimate is given by $\gamma^i$.
Because of the unbiasedness of the MVU/MC estimator the MSE is given by:
\begin{align*}
 \mse[\bar V_1] = \expect[\bar V_1^2] - V_1^2 = (1-p) \sum_{i=0}^\infty \gamma^{2i} p^i - 
\frac{(1-p)^2}{(1-\gamma p)^2} 
= \frac{p (1-p)}{(1-\gamma p)^2 (1-\gamma^2p)} (1-\gamma)^2,  
\end{align*}
where $\bar V_1$ denotes the MVU/MC estimator. For the MSE of the ML estimator $\bar{ \bar V}_1$ we need to
calculate the first and the second moment. The first moment:
\begin{align*}
\expect[\bar{\bar V}_1] = (1-p) \sum_{i=0}^\infty \frac{1-\bar p}{1-\gamma \bar p} p^i
=(1-p) \sum_{i=0}^\infty \frac{1}{1+(1-\gamma)i} p^i.
\end{align*}
In the following, we chose $\gamma$  such that $(1-\gamma)^{-1} = m \in \mathbbm{N}$.
The sum can then be written as
\begin{align*}
\frac{m(1-p)}{p^{m}}  \sum_{i=0}^\infty \frac{1}{m+i} p^{m+i} = 
\frac{m(1-p)}{p^{m}}  \left(\sum_{i=1}^\infty \frac{p^i}{i}  - \sum_{i=1}^{m-1} \frac{p^i}{i}    \right) 
= \frac{m(1-p)}{p^{m}}  \left(\ln\frac{1}{1-p}  - \sum_{i=1}^{m-1} \frac{p^i}{i}  \right). 
\end{align*}
 The second moment:
\begin{align*}
\expect[\bar{ \bar V}_1^2] &= (1-p) \sum_{i=0}^\infty \frac{(1-\bar p)^2}{(1-\gamma \bar p)^2} p^i
=(1-p) \sum_{i=0}^\infty \frac{1}{(1+(1-\gamma)i)^2} p^i \\
&=\frac{(1-p) m^2}{p^m} \sum_{i=0}^\infty \frac{1}{(m +i)^2} p^{m+i} =
\frac{(1-p) m^2}{p^m} \left(\sum_{i=1}^\infty \frac{p^i}{i^2} - \sum_{i=1}^{m-1} \frac{p^i}{i^2}  \right).
\end{align*}
The infinite sum is called \textit{Spence function} or \textit{dilogarithm} and is denoted with $Li_2(p)$. 
Using these terms one can derive the MSE:
\begin{align*}
\frac{(1-p)^2 m^2}{p^m (m (1-p) + p)} \Biggl( 
   &\frac{m(1-p) + p}{(1-p)} \left(Li_2(p) - \sum_{i=1}^{m-1}\frac{p^i}{i^2}\right) 
-  \frac{2}{m(1-p)}\left(\ln\frac{1}{1-p} - \sum_{i=1}^{m-1}\frac{p^i}{i} \right) \\
&+ \frac{p^m}{(m(1-p) +p)}\Biggr). 
\end{align*}
For $\gamma=p=1/2$ the MSE of the MVU/MC estimator is $0.127$ and $0.072$ for the ML estimator.
Contrary, for $p= 0.99$  the MSE of the MVU/MC estimator is $0.0129$ and $0.0219$ for the ML estimator.

\bibliographystyle{plainnat}
\bibliography{MDP}

\end{document}